\documentclass[manuscript,screen]{acmart}

\usepackage{booktabs}
\usepackage{wrapfig}
\usepackage{rotating}
\usepackage{pdflscape}
\usepackage{afterpage}

\usepackage[utf8]{inputenc}
\usepackage{array}
\usepackage{makecell}

\AtBeginDocument{%
  \providecommand\BibTeX{{%
    \normalfont B\kern-0.5em{\scshape i\kern-0.25em b}\kern-0.8em\TeX}}}

\setcopyright{acmcopyright}
\copyrightyear{2018}
\acmYear{2018}
\acmDOI{10.1145/1122445.1122456}

\acmConference[Woodstock '18]{Woodstock '18: ACM Symposium on Neural
  Gaze Detection}{June 03--05, 2018}{Woodstock, NY}
\acmBooktitle{Woodstock '18: ACM Symposium on Neural Gaze Detection,
  June 03--05, 2018, Woodstock, NY}
\acmPrice{15.00}
\acmISBN{978-1-4503-XXXX-X/18/06}

\begin{document}

\title{Camera Measurement of Physiological Vital Signs}

\author{Daniel McDuff}
\affiliation{%
  \institution{Microsoft Research}
  \city{Redmond}
  \country{USA}}
\email{damcduff@microsoft.com}

\begin{abstract}
The need for remote tools for healthcare monitoring has never been more apparent. Camera measurement of vital signs leverages imaging devices to compute physiological changes by analyzing images of the human body. Building on advances in optics, machine learning, computer vision and medicine these techniques have progressed significantly since the invention of digital cameras. This paper presents a comprehensive survey of camera measurement of physiological vital signs, describing they vital signs that can be measured and the computational techniques for doing so. I cover both clinical and non-clinical applications and the challenges that need to be overcome for these applications to advance from proofs-of-concept. Finally, I describe the current resources (datasets and code) available to the research community and provide a comprehensive webpage (\url{https://cameravitals.github.io/}) with links to these resource and a categorized list of all the papers referenced in this article.
\end{abstract}

\ccsdesc[500]{Human-centered computing~Ubiquitous and mobile computing}
\ccsdesc[500]{Applied computing~Bioinformatics}
\ccsdesc[500]{Computing methodologies~Computer vision}

\keywords{physiology, signal processing, machine learning, thermal imaging}

\maketitle

\section{Introduction}

\begin{wrapfigure}{L}{0.5\textwidth}
\centering
  \includegraphics[width=0.5\textwidth]{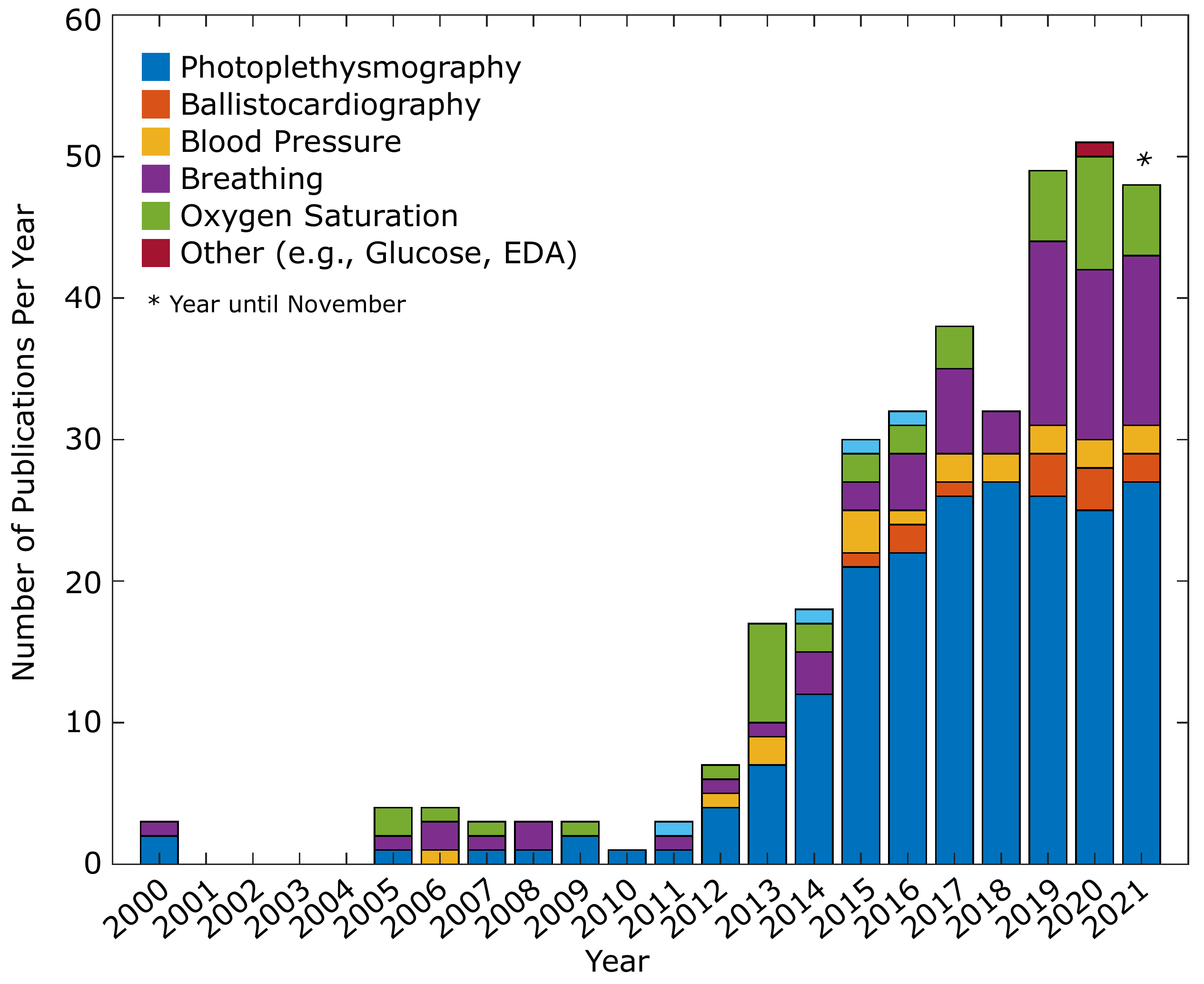}
  \caption{The number of publications per year indexed on PubMed.gov on contactless camera measurement of physiological vital signs.}
  \label{fig:no_papers}
\end{wrapfigure}

Camera measurement of vital signs has emerged has a vibrant field within computer vision and computational photography. This work combines expertise from these domains and those of signal processing, machine learning, biomedical engineering, optics and medicine, to create technologies that enable scalable and accessible physiological monitoring. The field has grown rapidly in the past 20 years, with papers published at an exponential growth rate (see Fig.~\ref{fig:no_papers} for an example). Using cameras for non-contact measurement has several distinct advantages and applications in a range of contexts. In telehealth, remote measurement of vital signs is an important tool in assessment and diagnosis, and cameras are a ubiquitous form of sensor available on almost every digital communications device (e.g., cellphone, PC, etc.). In inpatient ICU care, remote measurement can help protect patients and physicians while also creating a more comfortable experience for those receiving care, whether it be a baby~\cite{aarts2013non} or an adult~\cite{tarassenko2014non}. Less invasive sensing can help patients to sleep and eat while still being monitored. In low resource settings cameras are a cost effective and widely available form of sensor that can be easily transported and used for opportunistic measurement. Camera measurement of vitals could turn billions of devices with webcams into instruments for healthcare. 

However, camera physiological measurement presents numerous challenges that must be overcome before this potential can be fully realized. Physiological changes are often subtle, individually and contextually variable, and can easily be obscured by clothing, hair and makeup. Other changes in a video, such as from motion and illumination, can swamp the small pixel variations that contain cardiac and pulmonary information. Last, but by no means least, are the serious ethical and privacy implications of non-contact measurement. 

This article presents a survey of the field from foundations to state-of-the-art computational methods, discusses the applications of these tools and highlights challenges and opportunities for the research community. In this survey I focus on technologies that use visual spectrum, near infra-red (NIR) and far infra-red (FIR or thermal) cameras. These are all non-ionizing regions of the electromagnetic spectrum, making imaging safe for extended periods of time and in many cases possible without an active or dedicated light source. 
To accompany this survey, a website has been prepared with all the referenced papers categorized with key words and links to open source code repositories and datasets.

In my search of the literature, I used the following key words: `remote', `imaging', `non-contact', `camera', `video', `physiology',
`photoplethysmography', `rppg', `ippg'/`ppgi', `ballistocardiography', `respiration', `breathing', `pulse', `blood pressure', `electrodermal activity', `oxygen saturation', `glucose'. I primarily searched on Google Scholar, Microsoft Academic and PubMed. As an example, via PubMed I found over 215 papers on camera photoplethysmography and ballistocardiograpy that have been published in the past five years, an increase from approximately 60 in the previous five years~\cite{mcduff2015survey}. Across all terms I found over 350 papers published on camera physiological measurement on PubMed alone.\footnote{Based on results from https://pubmed.ncbi.nlm.nih.gov/}
There exist other surveys of camera methods~\cite{mcduff2015survey,sun2015photoplethysmography,chen2018video,shao2020noncontact,ni2021review} and some complementary comparative studies~\cite{wang2018comparative,ni2021review}. However, some of these only focus on signal processing methods and were published before supervised learning and deep learning came to the fore~\cite{mcduff2015survey,sun2015photoplethysmography} and others focus only on deep learning methods~\cite{ni2021review}, while others do not comprehensively cover the topic from foundations to computational methods to applications~\cite{wang2018comparative,shao2020noncontact}. I argue that given the significant advances in the research community in recent years and the interest in these tools given the growth of telehealth platforms, that a systematic survey of the field is warranted. A thorough survey of the literature will help to establish the current state-of-the-art, synthesize insights from different approaches and solidify key challenges for the community to solve.

\section{Foundations}

\begin{figure}
\centering
  \includegraphics[width=\textwidth]{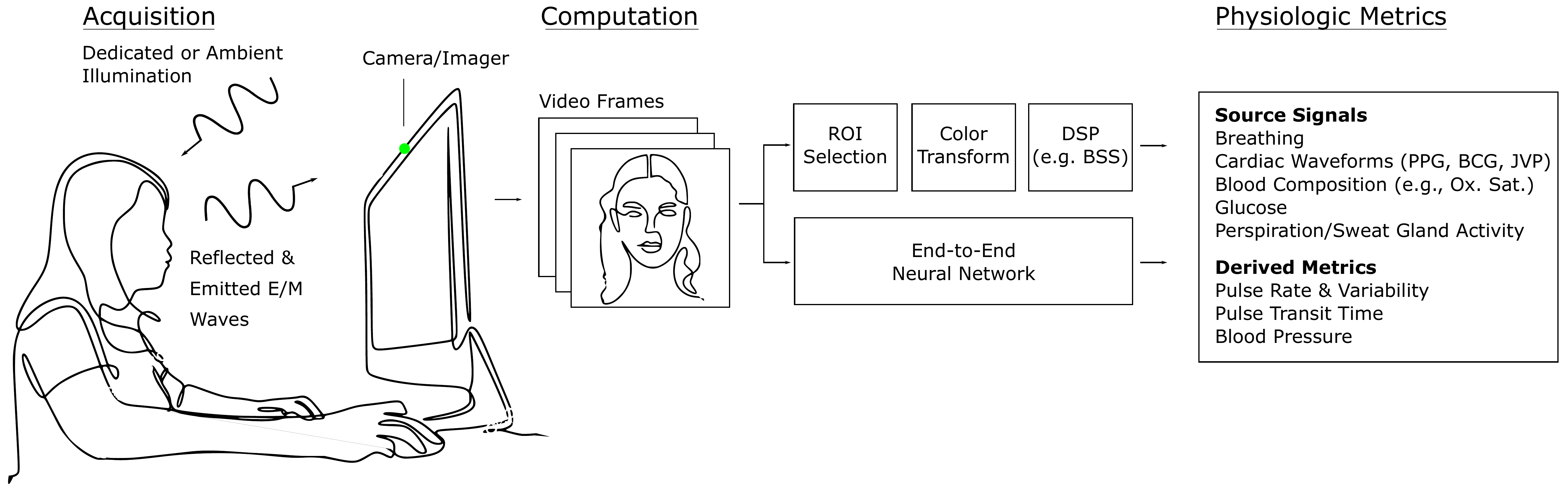}
  \caption{Camera measurement of vital signs has emerged has a vibrant field within computer vision and computational photography. Computational methods can be used to recover a range of physiological measures from imaging in the visual, near and far infra-red frequency ranges.}
  \label{fig:summary}
\end{figure}

\begin{wrapfigure}{L}{0.7\textwidth}
\centering
  \includegraphics[width=0.7\textwidth]{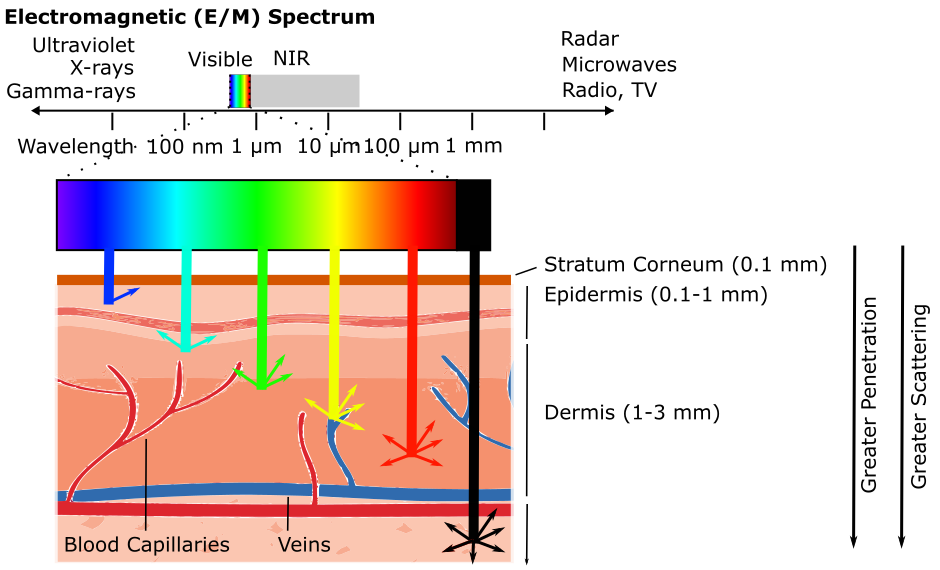}
  \caption{Camera physiological sensing uses non-ionizing E/M wavelengths to measure vital signs. As E/M wavelengths increase, the depth at which they penetrate the skin also increases; however, so does the amount of scattering that occurs.}
  \label{fig:spectrum}
\end{wrapfigure}

The advent of digital cameras created new opportunities for computation analysis of human bodies and physiology. Blazek, Wu and Hoelscher~\cite{blazek2000near} proposed the first imaging system for measuring cardiopulmonary signals. This computer-based CCD near-infrared (NIR) imaging system provided evidence that peripheral blood volume could be measured without contact using an imager. Shortly after this a similar approach was demonstrated using a visual band (RGB) camera~\cite{wu2000photoplethysmography}, devices that are considerably more ubiquitous than NIR cameras. Successful replications of this work cemented the concept~\cite{huelsbusch2002contactless,takano2007heart,garbey2007contact,verkruysse2008remote} and led to the growth of a new field of non-contact camera physiological measurement. Figure~\ref{fig:summary} illustrated the typical imaging pipeline for these system. As E/M wavelengths increase, the depth at which they penetrate the skin also increases (see Figure~\ref{fig:spectrum}); however, so does the amount of scattering that occurs. Depending on the signal of interest there is a trade-off between how much light is absorbed by the body and how much is reflected. Oxygenated blood, deoxygenated blood and skin tissue all have different absorption characteristics. Fortunately, within the visual bands close to 500-600nm in the ``green'' color range, there is a good trade-off between light penetration depth and hemoglobin absorption, making ubiquitously available RGB cameras a valuable tool for measuring cardiac signals via photoplethysmography. Camera technology has also been improving rapidly in quality due to investment from cellphone and other smart device manufacturers. Increases in sensor resolution and frame rate and reductions in sensor noise mean that subtle changes in motion due to pulmonary and cardiac activity can be captured. Work on body motion analysis from video, has found that to be a rich source of physiological information. enabling the recovery of breathing~\cite{tarassenko2014non} and cardiac signals~\cite{balakrishnan2013detecting}. These methods do not require light to penetrate the skin but rather use optical flow and other motion tracking methods to measure, usually very small, motions.

\subsection{Optical Model}

Optical models serve as a principled foundation for designing computational methods for camera physiological measurement.
For modeling lighting, imagers, and physiology, previous works have used the Lambert-Beer law (LBL) \cite{lam2015robust,Xu2014a} and Shafer's Dichromatic Reflection Model (DRM)~\cite{wang2016algorithmic,chen2018deepphys,liu2020multi}. As an example, let us take the DRM as our basis. 
Via analysis of video pixels we aim to capture both spatial and temporal changes and how these relate to multiple physiological processes. We start with the RGB values captured by the cameras as given by:
\begin{equation} \label{eq:1}
	\pmb{C}_k(t)=I(t) \cdot (\pmb{v}_s(t)+\pmb{v}_d(t))+\pmb{v}_n(t)
\end{equation}
where $I(t)$ is the luminance intensity level, modulated by the specular reflection $\pmb{v}_s(t)$ and the diffuse reflection $\pmb{v}_d(t)$. The quantization noise of the camera sensor is captured by $\pmb{v}_n(t)$. $I(t)$ can be decomposed into two parts $\pmb{v}_s(t)$ and $\pmb{v}_d(t)$: respectively~\cite{wang2017algorithmic}:
\begin{equation} \label{eq:2}
	\pmb{v}_d(t) = \pmb{u}_d \cdot d_0 + \pmb{u}_p \cdot p(t)  
\end{equation}

$\pmb{u}_d$ is the skin-tissue unit color vector; $d_0$ is the reflection strength which is stationary; $\pmb{u}_p$ is the relative pulsatile strength caused by the hemoglobin and melanin absorption and $p(t)$ represents the underlying physiological signals of interest.
\begin{equation} \label{eq:3}
	\pmb{v}_s(t) = \pmb{u}_s \cdot (s_0+\Phi(m(t),p(t))) 
\end{equation}

where $\pmb{u}_s$ is the unit color vector of the light source spectrum; $s_0$ and $\Phi(m(t),p(t))$ denote the stationary and varying parts of specular reflections; $m(t)$ denotes all the non-physiological variations such as changes in the illumination, head rotation, and facial expressions.  
\begin{equation} \label{eq:4}
	I(t) = I_0 \cdot (1+\Psi(m(t),p(t))) 
\end{equation}

where $I_0$ is the stationary component of the luminance, and $I_0\cdot\Psi(m(t),p(t))$ is the intensity variation as captured by the camera. As in \citep{chen2018deepphys} we can disregard products of time-varying components as they are relatively small, giving:
\begin{multline} \label{eq:7}
	\pmb{C}_k(t)\approx \pmb{u}_c \cdot I_0 \cdot c_0+\pmb{u}_c \cdot I_0 \cdot c_0 \cdot \Psi(m(t),p(t)) +
	\pmb{u}_s \cdot I_0 \cdot \Phi(m(t),p(t))+\pmb{u}_p \cdot I_0 \cdot p(t)+\pmb{v}_n(t)
\end{multline}
Pulse and breathing signals are in fact not independent \citep{liu2020multi}. As an example, the PPG signal captures a complex combination of both pulse and breathing information. Specifically, both the specular and diffuse reflections are influenced by related physiological processes. Respiratory sinus arrhythmias (RSA) is one example of this, RSA describes the rhythmical fluctuations in heart periods at the breathing frequency \citep{berntson1993respiratory}. Another example is that breathing and cardiac pulse signals both cause observable motions of the body. We can say that the physiological process \textit{p(t)} is a complex combination of the photoplethysmographic \textit{ppg(t)}, ballistocardiographic, \textit{bcg(t)}, and the breathing wave, \textit{r(t)}. Thus, $p(t) = \Theta(ppg(t),bcg(t),r(t))$ and the following equation gives a more accurate representation of the underlying process:
\begin{multline} \label{eq:8}
	\pmb{C}_k(t)\approx \pmb{u}_c \cdot I_0 \cdot c_0+\pmb{u}_c \cdot I_0 \cdot c_0 \cdot \Psi(m(t),\Theta(ppg(t),bcg(t),r(t))) +
	\pmb{u}_s \cdot I_0 \cdot \Phi(m(t),\Theta(ppg(t),bcg(t),r(t)))+\pmb{u}_p \cdot I_0 \cdot p(t)+\pmb{v}_n(t)
\end{multline}

There are alternatives to this optical model and it does not capture all physiological changes.  For example, around the neck the Jugular Venous Pulse (JVP) would be observed. However, this example does provides a basis or foundation for thinking about the design of computational methods that separate the source signals of interest from noise.

\section{Hardware}

\subsection{Visual Spectrum (RGB and Grayscale) Cameras}

\begin{figure}
\centering
  \includegraphics[width=\textwidth]{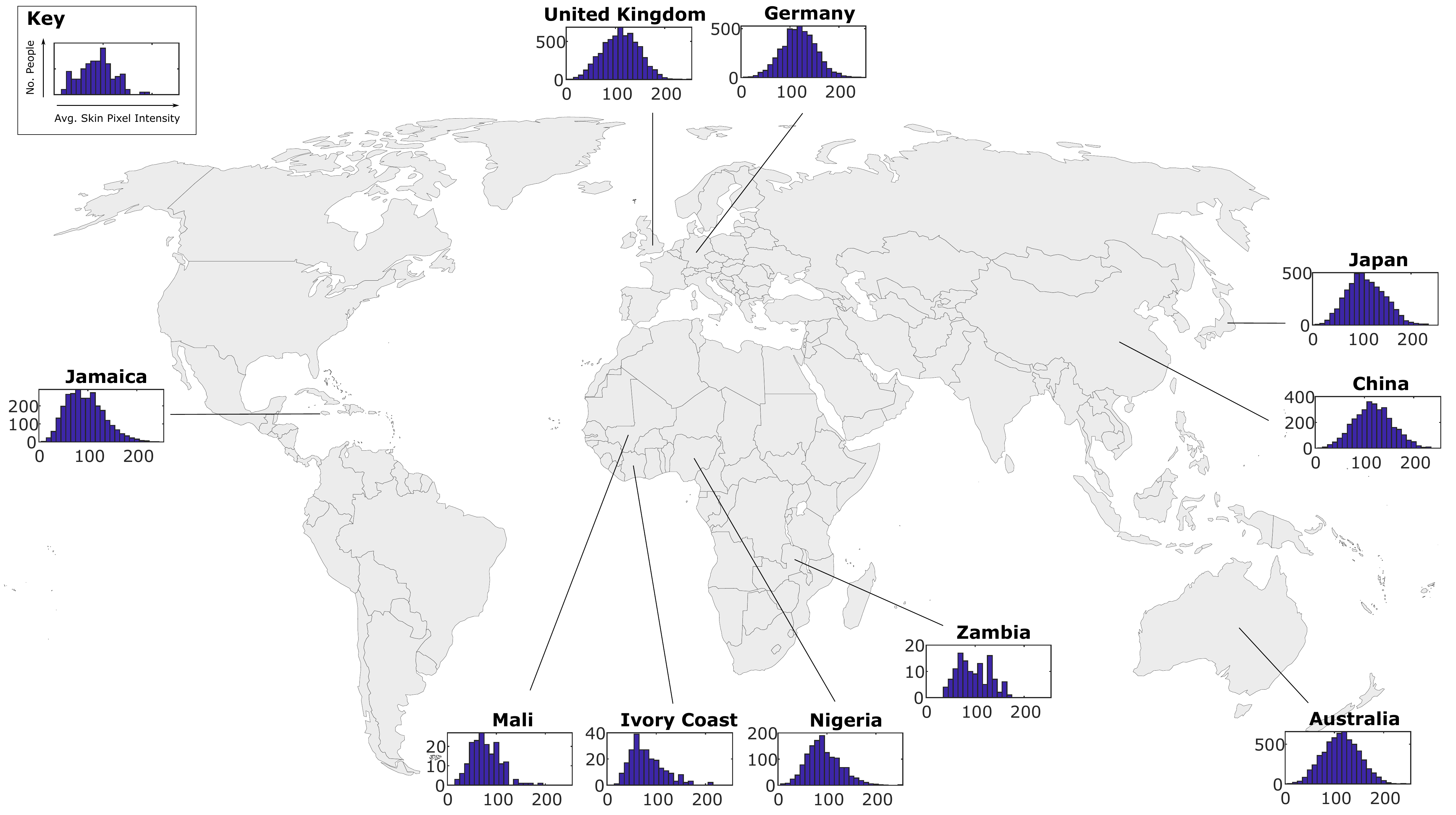}
  \caption{Histograms of average face skin pixel intensity for RGB photographs of people from several countries around the world. Notice how for Western and East Asian countries average face skin pixels tend to be distributed in the middle of the pixel range. Whereas for people from African or Caribbean countries these distributions are skewed heavily towards zero. The fact that camera have been optimized to capture lighter skin types has an impact on the performance of camera physiological measurement.}
  \label{fig:skinpixel}
\end{figure}

The visual light spectrum covers frequencies from 380 to 700 nanometers (nm). By far the most ubiquitous form of imager is the RGB camera. RBG imagers include webcams, cell phone cameras and digital photography cameras (e.g., DSLRs). Cameras are even now included on some smart TV's, in-home smart devices and doorbells, refrigerators and mirrors. These devices are typically optimized for visual clarity, creating images and videos that are clear to the human eye. Furthermore, they have often been optimized for affluent, western and Asian consumers, thus capturing lighter skin types more effectively then darker skin types\footnote{https://petapixel.com/2015/09/19/heres-a-look-at-how-color-film-was-originally-biased-toward-white-people/}. Figure~\ref{fig:skinpixel} shows the distribution of skin pixels for people from several countries around the world. Notice how the histogram of skin pixel values for those from Western countries (i.e., UK, Germany, Australia) fall close to the middle of the 0-255 pixel range with a Gaussian or normal distribution, whereas for those from African countries (Mali, Nigeria, Ivory Coast, etc.) or the Caribbean (Jamaica) the pixel values are skewed closer to 0. Signal-to-noise ratios are generally higher the lower pixel intensities become and saturation is more likely to occur for those subjects which would cause the changes in a video due to physiological variations to be lost. To compound this, face detection algorithms~\cite{mcduff2019characterizing} and similar tools~\cite{buolamwini2018gender} that are often used in camera physiological measurement pipelines often have biases.
As such, biases in the performance of physiological measurement using cameras not only stems from the optimization criteria, models and training data that are used but also from the hardware. I will discuss attempts to characterize and correct these biases in Section~\ref{sec:fairness}. However, it should be noted that little work has attempted to address disparities in performance resulting from hardware.

Assuming uniform illumination (e.g., broad spectrum/white light) the maximum signal-to-noise ratio for the blood volume pulse is at approximately 570nm \cite{blackford2018remote}, this is the frequency at which the the absorption of hemoglobin is greatest. However, if the illumination was particularly strong at another frequency, this could change. 
Surprisingly, measurements with RGB cameras can be made with reasonable precision up to 50 meters from the subject~\cite{blackford2017measurements}, which highlights not only the potential for this technology in remote measurement but also the potential for it to be used for covert surveillance and other troubling applications.  I will discuss the broader impacts of camera physiological measurement in Section~\ref{sec:broader_impacts}.

\subsection{Near Infrared Cameras}

Near-infrared (NIR) cameras sense light with wavelengths from 700nm to 1000nm. Human eyes are not sensitive to this wavelength range and therefore imaging systems can be designed with dedicated active light sources without interfering with human vision. However, hemoglobin absorption is weaker in this band compared to that of visible light and the PPG (or blood absorption signal) will typically have a lower signal-to-noise ratio. Although, a systematic comparison of RGB and IR camera measurement of all possible physiological parameters (including motion-based signals such as BCG and breathing) has not been performed to our knowledge. Such an analysis would be a valuable contribution. The fact that IR cameras cannot image colors in the visible range may limit the precision with which changes in motion can be measured. But it is also reasonable to think that motion-based measurement would be less affected than reflectance-based PPG measurement that captures blood volume.

NIR cameras have the distinct advantage of being able to image in low light conditions. Due of this, they are particularly suited for measuring physiological parameters during sleep and at night time. As sleep studies, baby monitoring and driving are all examples of applications that could benefit from non contact measurement~\cite{amelard2018non,vogels2018fully,van2020camera,scebba2020multispectral}, NIR cameras are attractive. Several studies have identified the possibility of detecting the effects of sleep apnea events on both the PPG and breathing signals using these devices.

Another reason that NIR cameras may be attractive for certain applications is that the photometric properties of the skin are not as strong between 700-1000nm and therefore differences in performance by skin type might be lower. However, I am not aware of any empirical work that provides a systematic quantitative analysis of this, another example of a potentially valuable contribution to the research literature.

\subsection{Thermal Cameras}
Thermal cameras measure far infrared (FIR) signals, covering wavelengths from 2000nm (2 microns) to about 14000nm (14 microns). At these wavelengths, most objects even in ambient temperature radiate heat in the form of thermal emissions. As humans are emissive sources, there is no need for ``illumination'', either ambient or controlled. Thermal sensors have certain capabilities that are beyond those of RGB and NIR cameras. The most obvious being that FIR signals allow for measurement of human body temperature without contact, something that is not possible with RGB and NIR cameras. However, there are other examples, imaging of sweat glands to measure transient perspiration and dermal responses is possible~\cite{shastri2012perinasal}. When activated, perspiration pores lower thermal emission, absorbing latent heat and they appear as colder in the thermal images. 

One challenge with thermal sensors is that in the FIR bands, sensor technologies are either based on micro-balometers (black-body adsorption) or materials such as Indium Gallium Arsenide (In-Ga-As) that are much more expensive than Silicon, running at several hundreds or even thousands of dollars rather than tens of dollars which is the case with many RGB cameras.
As a consequence, thermal sensors are typically lower resolution, lower SNR and much more expensive than RGB or NIR sensors. 
Some thermal sensors also require cooling, which means that they consume considerably more power than RGB and NIR imagers. Lower cost and more portable thermal cameras have been developed in recent years, these devices initially had very low spatial resolution but that has also changed. Some ``off-the-shelf'' thermal cameras are now available for under \$100.

\subsection{Multi- and Hyper-Spectral Cameras}

Studies with RGB, NIR and FIR cameras are the most common in the literature. However, the design and use of other multi- and hyper-spectral cameras has been explored. As previously described, certain wavebands are better for PPG-derived pulse rate and breathing rate measurement than others and it has been shown imaging multiple image bands can improve the robustness of physiological signals~\cite{mcduff2014improvements,spigulis2017multispectral}.
Given these results, one might reason that spectral bands common in most digital imagers may not be optimal in terms of resolution, range, and sensitivity for physiological measurement, especially when it comes to measuring absolute changes in blood composition, transient perspiration or other physiological meaningful signals.

While there is potential for performance gains using hyperspectral imaging, or imaging spectroscopy, the availability of this type of novel sensors is limited in practice due to their cost, mismatched temporal resolution (push-broom vs. snapshot, or global shutter, image acquisition) and lack of ubiquity (in comparison to standard digital cameras in the visible and near-infrared ranges).
Despite these limitations, investigations into multiband imaging, as well as imaging outside of the visible and near-infrared ranges, could further prove fruitful and there are several applications of camera physiological measurement in which customized hardware might be appropriate. As an example, a five-band visible band camera, with cyan and orange pixels present (in addition to standard RGB) produced results that outperformed RGB which using the waveband combinations orange, green, and cyan~\cite{mcduff2014improvements}. Future work may seek to explore multiband, and potentially multi-imager, applications that leverage spectrally-tuned approaches to integrating multiple wavebands. For example, this could be achieved through the use of specifically tuned optical filters across a small, spatially redundant array of both visible and infrared imagers.

Fusing predictions rather than building a new hardware from, much of the research into multispectral imaging uses more than one camera, each with sensitivity in a different range (e.g., monochrome cameras with filters or NIR and FIR cameras). The predictions from these can then be fused~\cite{scebba2020multispectral} or combined at a feature~\cite{he2021spatiotemporal} level to make estimates. He et al.~\cite{he2021spatiotemporal} use a calibration process to decompose RGB images into multispectral cubes thereby creating a larger feature space for performing signal separation.

\section{Physiologic Measures}

\subsection{Cardiac Measurement}

Using cameras there are several methods of cardiac pulse measurement that have been developed, via photoplethysmography (PPG), ballistocardiography (BCG) and the Jugular Venous Pulse (JVP). These are all well established non-invasive instruments, but have traditionally been measured using customized contact hardware. I will discuss each of these in detail below, including the pros and cons offered by each approach and how they can be fused together. It should be noted that many of the same techniques for measuring PPG also apply to BCG and JVP. Therefore, in our discussion of algorithmic approaches to measurement almost all can be considered as applying to all of the signals.

\textbf{Photoplethysmography.} Photoplethysmography involves measurement of light transmitted through, or reflected from, the skin and captures changes in the subsurface blood volume pulse~\cite{allen2007photoplethysmography}. Non-contact camera imaging almost always leverages reflectance PPG as imaging is generally performed from the head or other region of the body through which light will not transmit (unlike the ear lobe or fingertip) and there is usually no dedicated illumination source bright enough to transmit through even thin body parts. In videos, the PPG manifests as very subtle pixel color changes of the skin. Sometimes these changes can be ``sub-pixel'' meaning that the light changes are less than one bit change per pixel per frame. However, by averaging a region of pixels together a continuous signal can be recovered. The simplest way to think of this is as the camera as a collection of noisy pixel sensors that when aggregated together reveal the desired continuous signal.
There are clinical applications for PPG measurement as the signal contains information about the health state and risk of cardiovascular diseases~\citep{elgendi_use_2019, reisner_utility_2008, pereira_photoplethysmography_2020}. It could be argued that the PPG has been underutilized for clinical applications and as more methods are developed that make measurement easier, more convenient and more accurate we will see greater impact. 

\textbf{Ballistocardiography.} Ballistocardiography involves measurement of the mechanical motion of the body due to the cardiac pulse~\cite{starr1939studies}. Although the PPG and BCG are measured via different mechanisms they can often both be present in the same video because typically skin regions that feature PPG information also exhibit BCG motions. This can be used to help improve the estimates of downstream metrics such as pulse rate~\cite{lomaliza2020combining}.
However, it can also mean that the PPG and BCG signals are difficult to separate in a video~\cite{moco2015ballistocardiographic}. The BCG signal can be measured from video by tracking fiducial/landmark points or via optical flow to capture the subtle motions of the body resulting from the cardiac pulse~\cite{balakrishnan2013detecting,shao2016simultaneous}. The BCG provides complementary information to the PPG signal and could be used to help derive metrics related to pulse wave velocity or pulse transit time~\cite{shao2016simultaneous}. However, camera-based BCG measurement methods are highly sensitive to other body motions making it difficult to design practical applications that leverage this information, except in highly controlled contexts. One advantage of the BCG measurement is that it is not dependent on the presence of skin and therefore can be used to recover cardiac signals from the back of the head or a body part covered in clothing~\cite{balakrishnan2013detecting}.

\textbf{Jugular Venous Pulse.} The Jugular Venous Pulse is a diagnostic tool used to assess cardiac health. The jugular vein is an extension of the heart's right atrium and changes in atrial pressure can be reflected in the jugular waveform. The JVP is measured by analyzing the motion of the neck, just below the chin. Specifically, distortions in the JVP waveform morphology can provide non-invasive insight into cardiac function~\cite{amelard2017non}.  Using camera methods, primarily capturing motion or optical flow, it has been shown that the JVP can be measured optically~\cite{amelard2018non,garcia2020extracting}. Aiding clinicians in the observation of this signal can help with bedside examinations~\cite{abnousi2019novel}. Again, the JVP contains complementary information to the PPG and BCG waves, meaning that combining these signals offers opportunities for additional insights into cardiac function.  The fact that all three can be captured from the same sensor data, a video, reduces the complexity of measurement and synchronization of observations. I anticipate researchers will leverage combinations of these signals to greater effect in future.


\textbf{Pulse Rate.}
The pulse rate (PR) is the dominant frequency within the PPG, BCG and JVP waveforms and is typically the simplest information to derive.  If the periodic peaks corresponding to the heart beats are not observable in the waveform, then other metrics will likely be very difficult to measure. I am careful to use the term pulse rate here; however, this will be very similar to heart rate (HR) in most cases. Looking for the periodic systolic peaks in the video cardiac waveforms is one way to assess their signal quality. Essentially, the BVP signal-to-noise ratio (SNR) proposed by De Haan~\cite{de2014improved} and frequently used in evaluating video PPG measurement captures that fact. Most camera methods concerned with cardiac measurement have evaluated performance in terms of average or instantaneous pulse rate measurement. While this is a logical place to start, moving forward I hope that in addition to average pulse rate measurement increasing emphasis is placed on other metrics too.

\textbf{Pulse Rate Variability.} Pulse rate variability (PRV) captures the changes in pulse rate over time and is a commonly used measure of autonomic nervous system (ANS) activity. Heart rate variability (HRV) is closely related to PRV and can in many cases be very similar~\cite{mcduff2014improvements}. The two branches of the ANS are the sympathetic nervous system (SNS) and parasympathetic nervous system (PNS) which dynamically control the beat-to-beat differences of the heart. The HRV low frequency (LF) component is modulated by baroreflex activity and contains both sympathetic and parasympathetic activity~\cite{akselrod1981power}. The high frequency (HF) component reflects parasympathetic influence on the heart, it is connected to respiratory sinus arrhythmia (RSA). An estimate of sympathetic modulation (the sympatho/vagal balance) can be made by considering the LF/HF power ratio.  PRV can be computed in several ways, but generally requires detecting the pulse inter-beat intervals. The computation can be quite sensitive to the precision of the inter-beat measurement which presents challenges for computing PRV derived metrics in many applications. Increasingly, work in camera physiological measurement is being evaluated on the performance of inter-beat measurement rather than average PR\cite{revanur2021first}, this is an encouraging trend as it sets a higher bar for algorithmic performance. End-to-end networks could be used to predict peak timings directly from video, rather than recovering a waveform and performing peak detection. Sequence-to-sequence models might be quite effective at this task; however, a rigorous evaluation of such an approach for PRV measurement has not been performed.

\textbf{Breathing.}
By leveraging RSA, breathing or breathing rates can be derived from the PPG, BCG or JVP signals by analyzing the HF components of the PRV~\cite{poh2010advancements}. However, this is not a perfect method as some irregular breathing patterns may not be clear within the heart rate variability and RSA can fluctuate in intensity. When someone is under stress it can be weaker than when they are at rest. Furthermore, since pulse rate variability itself is difficult to derive from a noisy cardiac signal, trying to measure breathing rate via the cardiac pulse variability can be unreliable. However, the principal does help motivate why multi-task modeling of physiological signals may be a promising direction~\cite{liu2020multi}. 

\textbf{Pulse Transit Time.}
There are several attractive properties for imaging systems in physiological measurement. One is that imaging systems can measure signals spatially, as well as temporally. Another is that cardiac pulse measurements can be made via multiple modalities (e.g., PPG, BCG and JVP measured simultaneously). Both of these properties enable some promising opportunities for measuring pulse wave velocity (PWV) or pulse transit time (PTT) (i.e., the time it takes for the cardiac pulse to reach a specific part of the body and therefore the velocity of that wave). Researchers have proposed two methods for doing so using imaging systems. The first involves measurement of the PPG signal at two locations on the body from the same video sequence. Shao et al.~\cite{shao2014noncontact} show that pulse arrival times at the palm and the face can be measured and contrasted. Other work uses the time delay between different cardiac pulse wave (e.g., BCG and PPG) major (or systolic) peaks~\cite{shao2016simultaneous}.
Building on this work, it may be possible to measure more dense or continuous spatial variation across the body using imagers, rather than just two locations. Several papers have shown examples of such visualizations but have not validated this or compared it to downstream metrics such as PTT or blood pressure.

\textbf{Arrhythmia.} Cardiac arrhythmia, such as atrial fibrillation (AF), are a predictor of serious cardiac events. Over 30\% of cardioembolic strokes are directly attributable to AF~\cite{yan2018contact}. Although not all forms of arrhythmia may be detectable via all cardiac signals, AF is possible to identify from the PPG signal. Studies have compared measurement using mobile phone cameras imaging of the fingertip and the face~\cite{yan2018contact,poh2018diagnostic}. Premature Ventricular Contractions (PVC) are another form of arrhythmia that has been studied from contact measurements. Detection of PVC from the PPG and BCG waveforms is possible as summarized in~\cite{shao2020noncontact}. But there is no published camera-based measurement work to our knowledge. Qualitatively, the author has observed PVC in camera PPG data that were validated by ECG measurements.


\textbf{Morphological Features.}
The cardiac pulse ways have interesting morphological features. Distortion of the JVP can provide information about cardiac function~\cite{amelard2017non}.
In the PPG signal, each pulse wave features a systolic peak and diastolic peak separated by a dichotic notch or inflection. 
Fingertip analysis of PPG signals has revealed the promise of these features for downstream assessments~\cite{elgendi2012analysis}.
Using these features, metrics such as the left ventricle ejection time (the time between the systolic foot and the dichotic notch) can be derived. For assessing cardiac health, morphological features could be more important than heart rate or heart rate variability.

However, accurately measuring these subtle waveform dynamics is non-trivial.  For example, the dicrotic notch may only manifest as an inflection in the raw PPG wave; however, in the second derivative this inflection is a maxima. Computing the second derivative, or acceleration PPG, can be a useful tool for extracting waveform features. The second derivative of the PPG signal can be used as an indicator of arterial stiffness - which itself is an indicator of cardiac disease~\cite{inoue_second_2017}, similar information contained with the wave can be used to estimate vascular aging~\cite{takazawa_assessment_1998}, which was higher in subjects with a history of diabetes mellitus, hypertension, hypercholesterolemia, and ischemic heart disease compared to age-matched subjects without.

Under controlled conditions camera algorithms can measure subtle morphological features. McDuff et al.~\cite{mcduff2014remote} evaluated measurement of systolic-diastolic peak-to-peak time (SD-PPT) and Hill et al.~\cite{hill2021learning} evaluated left ventricle ejection time (LVET).
In the latter it was observed that optimizing for the second derivative error directly, rather computing it from a lower-order prediction, can improve the accuracy of that measure, presumably because the dynamics of the waveform morphology are more faithfully preserved.

\textbf{Blood Pressure.} Some of the metrics derived from camera physiological measurements are correlated with blood pressure.  However, currently there is little evidence that cameras could be used to directly measure blood pressure. 
Morphological features in the PPG wave do contain some information about blood pressure. Using a network trained on contact sensor data and then fine-tuning that on rPPG signals Schrumpt et al.~\cite{schrumpf2021assessment} were able to show reasonable BP prediction.
Utilizing the spatial measurement opportunities presented by cameras, researchers have shown that non-contact PPG from the face and palm can be used to derive pulse transit time which has then been correlated with blood pressure~\cite{shao2014noncontact,jeong2016introducing}. However, these were relatively small studies. A larger study with over 1300 subjects found that pulse amplitude, pulse rate, pulse rate variability, pulse transit time, pulse shape, and pulse energy features extracted from non-contact PPG measurements could be used to predict systolic pressure and diastolic pressure with reasonable performance~\cite{luo2019smartphone}. While these are promising results, their study only features normotensive subjects and not hypertensive or hypotensive patients. Further work is needed to build confidence in the potential of camera measurement of blood pressure, but the opportunities that that would present are obvious, therefore I expect this to be an area of active research.

\subsection{Pulmonary Measurement}

There are many parallels in the methods used for pulmonary measurement as for cardiac measurement. Using cameras the most obvious method for measuring pulmonary activity is analyzing motion of the body primarily the torso, mouth and nostrils. In their simplest form these algorithms uses pixel averaging to capture changes in luminosity within a video over time. As with cardiac measurement, these naive methods can be improved by segmenting a region of interest rather than using the whole frame, but neither case will typically lead to robust estimates in the presence of other motions or lighting changes. To improve upon this, two forms of motion analysis have been proposed the first involving tracking fiducial, or landmark, points and the second involving measuring optical flow at a pixel level (i.e., dense flow). There are numerous examples up these approaches applied to camera breathing measurement from RGB~\cite{bartula2013camera,tarassenko2014non,chen2018deepphys,lorato2021towards}, NIR~\cite{bartula2013camera,chen2018deepphys} and FIR images~\cite{lewis2011novel,pereira2018noncontact,lorato2021towards}. The combination of modalities/sensors has also been explored~\cite{negishi2020contactless} and methods evaluated on more than one modality within the same study~\cite{lorato2021towards}. End-to-end supervised neural architectures have also been used. These neural architectures are trained in a similar fashion to cardiac measurement systems with pixels forming the input and a loss computed on the predicted breathing waveform~\cite{chen2018deepphys}.
Solutions that rely on sparse landmark points will be limited in their potential as inevitably additional information available within the video will be ignored.

\textbf{Breathing Rate.}
The breathing rate is the dominant frequency within the breathing waveform and, as with heart rate, is typically the simplest information to derive. Breathing rates of 12 to 20 breaths per minute are normal at rest.  Lower breathing rates may be observed during apnea events or exercises such as meditation. Higher breathing rates would generally be observed during physical exercise. It is not uncommon for successive breaths to vary considerably in duration from one another, which may mean that frequency domain analysis of the breathing waveform does not lead to one dominant peak. 

\textbf{Breathing Rate Variability.} Similar to cardiac signals, the variability of breathing rates can be a useful signal about how the body is functioning. Breathing rate variability has not been studied as much as pulse or heart rate variability. 

\textbf{Tidal Volume.}
Tidal volume is the amount of air that moves in or out of the lungs with each respiratory cycle. Measuring this signal involve not only the duration and depth of each breath but the volume of the chest~\cite{lewis2011novel}. This can be simplified as relative tidal volume, which requires only measuring the relative volume changes.

\subsection{Electrodermal Measurement}

Electrodermal activity is it change in conductance of the skin in response to sweat secretions. sweat glands are of the order of 0.05mm to 0.1mm and are not visible to the unaided eye nor can they typically be measured using RGB cameras. Thermal cameras are able to measure sweat gland activity via the changes in thermal emissions. Using this technique research has revealed how to measure changes in the diameter of the gland in the perinasal region, which can be used to measure transient perspiration~\cite{shastri2012perinasal}. Using RGB cameras it is certainly possible to measure correlates of electrodermal activity~\cite{bhamborae2020towards}. These could include BVP amplitude and vasomotion.  NIR cameras may under some conditions, be able to measure moisture on the surface of the body; however, there is little evidence at the moment that this would be effective at capturing a signal that correlates highly with electrodermal activity.

\subsection{Blood Oxygen Saturation} 

The composition of the blood can be measured using cameras with multiple frequency bands, one can think of this as a low frequency resolution form of spectroscopy. However, because of the broad frequency sensitivity of most RGB cameras calibration can be challenging. Oxygen saturation, or the ratio between oxygenated and deoxygenated hemoglobin is the most well studied.  In non-contact camera measurement preliminary studies have validated that oxygen saturation can be captures using RGB cameras~\cite{tarassenko2014non}. Another method measures the total blood concentration as a function of oxygenated and deoxygenated blood~\cite{nishidate2011noninvasive}. This method requires calibration using a known color reference.

\subsection{Glucose}

Given the significance for patients with diabetes, the non-invasive measurement of blood glucose levels is another attractive goal. However, unlike oxygen saturation the variations in light measured via reflectance methods due to changes in glucose may be very difficult to detect. According to modeling by Wang et al.~\cite{wang2019modeling} it is unlikely to detect the blood glucose based on either the DC or AC component of skin reflected light. Their model capture light in the visible to NIR range.
Nevertheless, advances in the spatial, temporal and sensitivity of imaging hardware plus additional color bands could still present opportunities for non-contact camera glucose measurement.

\section{Computational Approaches}
\label{sec:computational_approaches}

The use of ambient illumination means camera-based measurement is sensitive to environmental differences in the intensity and composition of the incident light. Camera sensor differences mean that hardware can differ in sensitivity across the frequency spectrum. Automatic camera controls can impact the image pixels before a physiologic processing pipeline (e.g., white balancing) and video compression codecs can further impact pixel values. People (the subjects) exhibit large individual differences in appearance (e.g., skin type, facial hair) and physiology (e.g, pulse dynamics). Finally, contextual differences mean that motions in a video at test time might be different from those seen in the training data.

\subsection{Signal Processing Methods.} 
In the context of video based physiological measurement traditional signal processing techniques have several advantages. They provide simple to implement and often computationally efficient algorithms for the measurement of the underlying physiological signals. They are often also easy to interpret and relatively transparent. Most signal processing methods do not require training data (i.e., are unsupervised), which contributes to their simplicity and interpretability. 

Early methods for PPG and breathing measurement leveraged spatial redundancy to cancel out camera quantization noise and recover the underlying waveform~\cite{takano2007heart,verkruysse2008remote,tarassenko2014non}. These methods work well on raw videos with limited body motion and homogeneous lighting conditions; however, the presence of motion (either from the camera or subject), illumination changes, video compression artifacts and other sources of noise can easily corrupt the measurements. To address this, researchers proposed using blind-source signal separation (BSS) techniques such as independent component analysis (ICA)~\cite{poh2010non,poh2010advancements} and principle component analysis (PCA)~\cite{lewandowska2012measuring,wedekind2017assessment}. These are simple unsupervised learning techniques that can recover demixing matrices (usually linear) and optimize for certain signal properties. Typically, the demixing is performed frequently (i.e., every 30 second time window) so that the algorithm can adapt to changes in the video over time. In the case of ICA this optimization is typically performed by maximizing the non-gaussianity of the recovered signals. BSS methods often work effectively at removing noise from the waveforms when it is small in amplitude or relatively periodic. However, they make naive assumptions about the properties of the underlying waveforms. Given that we have prior knowledge about the physical and optical properties of the material (skin) and the physiological waveform dynamics it is reasonable to think that we could leverage those to improve our signal estimates, indeed this is what has been shown.

Chromiance-based methods~\cite{CHROMdeHaan,de2014improved} are such an example, these are designed with the aim of eliminating specular reflections by using specifically tuned color differences. Building two orthogonal chrominance signals from the original RGB signals (specifically, X = R - G and Y = 0.5R + 0.5G - B) helps improve the PPG signal-to-noise ratio. Of course, these are specific to the measurement of absorption changes and not body motions.
Wang et al.~\cite{wang2017algorithmic} proposed another physically-grounded demixing approach based on defining a plane orthogonal to the color space of the skin (POS) which one of the most robust signal processing methods for PPG recovery.  Another physiologically-grounded approach used a physical synthetic skin model for learning demixing parameters using Monte Carlo methods~\cite{nishidate2011noninvasive}.  
Pilz~\cite{pilz2018local} used principles of local group invariance and then built upon this approach~\cite{pilz2019vector} to define a lower, or compressed, dimensional embedding of the pixel space that performed competitively for PPG signal recovery. One attractive property of these demixing and group invariance methods, is that they can be very fast to compute at test or run time. 


For motion-based signal recovery similar approaches have been applied, Balakrishnan, Durand and Guttag~\cite{balakrishnan2013detecting} used feature tracking to form a set of temporal signals and then PCA to recover the BCG signal. This method was adopted to compute the velocity and acceleration BCG signals in other work~\cite{shao2016simultaneous}. Hernandez et al.~\cite{hernandez2014bioglass} used a similar approach applied to ego-centric videos, where the landmark tracking was applied to objected in the environment rather than points on the head or body.

For breathing,similar signal processing methods have been adopted using PCA and ICA~\cite{jorge2018data}. Other related work used auto-regressive (AR) filters, averaging pixels as the first step and then performing pole selection from the AR filter model fit to the temporal pixel average signal~\cite{tarassenko2014non}. Still other method~\cite{bartula2013camera} averaged pixels one only one axis (vertical) to create a 1D representation, filtering that representation and then performing correlations of these vectors across frames within a video.  

Given the periodic nature of the cardiac pulse and breathing signals filtering can significantly improve the signal-to-noise ratio and downstream metrics. Many methods apply bandpass filtering using a Hamming window, others use methods such as Continuous Wavelet Filtering~\cite{bousefsaf2013continuous}. To fairly compare computational methods, it is vital to ensure that filtering parameters are kept constant; unfortunately, there are numerous cases in the published literature in which filter cut-offs, order and window types are not reported.  For a given dataset, results can be significantly improved by tuning filter parameters, but that does not capture the performance of the underlying signal recovery algorithm.

All these signal processing approaches have similar pitfalls. They often struggle to effectively separate noise from different sources and in most cases ignore a lot of spatial and color space information by aggressively averaging pixels early in the processing pipeline or computing the positions of a sparse set of spatial landmarks. It would seem that more complex temporal-spatial and colorspace representations would yield signals that more faithfully reflect the underlying physiological process, this is where supervised learning and deep neural models can offer advantages.

\subsection{Supervised Learning}

\textbf{Convolutional Models.}

Convolutional networks are the most common form of supervised learning used for camera physiological measurement. These networks learn representations using convolutional filters applied spatially or spatio-temporally to the input frames. DeepPhys~\cite{chen2018deepphys} was the first to propose a convolutional attention network (CAN) architecture trained using a combination of appearance frames and motion (difference) frames for physiological measurement. The two representations were processed by parallel branches with the appearance branch guiding the motion branch via a gated attention mechanism. The target signal was the first differential of the PPG wave. {\v{S}}petl{\'\i}k et al. also proposed a two part network, but in this case the networks were applied sequentially with an ``extractor'' network learning representations that were then input to an ``HR prediction'' network~\cite{vspetlik2018visual}. Loss was computed on the HR estimates. Liu et al.~\cite{liu2020multi} extended the CAN model to include multi-task prediction of both the PPG and breathing wave, thereby effectively halving the computational cost of using two networks with little reduction in accuracy.

For PPG estimation, spatial attention mechanisms essentially act as skin segmentation maps, perhaps learning to weight areas of skin with higher perfusion more heavily although this has only been validated qualitatively. Chaichulee et al.~\cite{chaichulee2019cardio} explicitly modeled skin segmentation in their network architecture before extracting the PPG and respiration signals. For breathing the skin region may or may not be the best source of information, as in many applications the chest may be the strongest source of breathing motions, but may be covered with clothing.

Using the ability of a convolutional network to perform video enhancement, essentially to remove noise, Yu et al.~\cite{yu2019remote} proposed a two stage process, the first is an encoder-decoder used to enhance the video, removing artifacts and noise, and the second is a PPG extraction network. Another method that has achieved strong results uses a different form of preprocessing. Niu et al.~\cite{niu2019robust,niu2020video} form spatio-temporal maps by computing average pixel intensities from different regions of the face and different color spaces (RGB and CYK). A convolutional network is then trained with these maps as input and the HR as the target. By preprocessing the signal in this way, the designer can incorporate prior knowledge about the spatial and color space properties of the desired signal. The trade-off is the additional computational and implementation costs that are incurred.

Given the characteristic morphology and periodicity of many physiological signals sequence learning (e.g., via an LSTM or RNN) can help remove noise from predicted waveforms~\cite{liu2018learning,yu2019remote,lee2020meta,nowara2021benefit,hill2021learning}. 
Yu et al.~\cite{yu2019remote} compared a 3D-CNN architecture with a 2D-CNN + RNN architecture finding that a 3D-CNN version was able to achieve superior PR prediction errors - suggesting that spatial-temporal modeling is more effective when information can be shared. Liu et al.~\cite{liu2020multi} found 3D-CNNs to be a good solution in terms of accuracy but with a large computational overhead.
Nowara et al.~\cite{nowara2021benefit} used the inverse of an attention mask to compute a noise estimate that was also provided as input to the sequence learning step, this noise prior helped to improve PPG estimates in the presence of motions.

Researchers have attempted to build multi task models that predict cardiac and pulmonary signals~\cite{liu2020multi}, but while there is certainly redundancy in the representations learned that can help reduce the computational demands of running multiple models in parallel, accuracy of measurement did not improve.

\textbf{Transformers.}
Transformers are becoming the architecture of choice for many computer vision tasks. They offer attractive trade offs between computation and scalability with training sets. By avoiding computationally expensive convolution operations and leveraging attention mechanisms heavily they are able to often provide a good balance between accuracy and efficiency. Preliminary work in camera physiological measurement has shown that these architectures are competitive with the state of the art convolutional networks~\cite{liu2021efficientphys}, but it is unclear whether with larger datasets it will be possible to exceed the performance of those convolutional baselines.  Transformers have also been applied with some success for breathing measurement~\cite{kwasniewska2021improving}, but both of these works are early investigations and more experimentation is needed.

\textbf{Support Vector Machines.}
A small number of other supervised methods have been proposed, for example, using support vector machines (SVM)~\cite{osman2015supervised}. However, they are relatively few and far between. With the dominance of neural models and importance of attention mechanisms in this task, we might infer that these other methods would be unlikely to exceed state-of-the-art performance. 

\subsection{Unsupervised Learning}

\textbf{Generative Adversarial Networks.}
Other methods have used generative adversarial networks to train models to generate realistic PPG waveforms. Pulse GAN~\cite{song2021pulsegan} (GAN stands for Generative Adversarial Network) is one such example, in which the authors used a chromiance signal as an intermediate representation during the training process. The Dual-GAN~\cite{lu2021dual} method involves segmentation of multiple facial regions of interest using a set of facial landmarks. These regions of interest are then spatially averaged and transformed into both RGB and YUV colorspaces. Using these data spatio-temporal maps (STMaps) are constructed which form the input to a convolutional network. This method produces strong results thanks to careful segmentation and the ability to leverage multiple color space representations. However these preprocessing steps are certainly nontrivial to implement and come at a significant computational cost~\cite{liu2021efficientphys}.

\textbf{Contrastive Learning.}
Training with unlabeled videos is highly attractive in a domain of camera physiological measurement as well synchronized datasets with videos and ground truth signals are difficult to obtain. Research has shown that training in an unsupervised fashion can be successful~\cite{liu2021metaphys,gideon2021way}. Contrastive learning is one tool that can be used for learning from unlabeled videos. Gideon et al.~\cite{gideon2021way} present a clever self-supervised contrastive learning approach in which they resample videos using video interpolation to create positive and negative pairs. Positive pairs have a matching HR and negative pairs have a different HR as a result of the resampling. This model can then be fine tuned in a supervised manner on a smaller data set. This approach achieved strong results obtaining the best performance on the Vision4Vitals challenge~\cite{gideon2021estimating}.

In other domains of computer vision pre-trained models have proven very powerful tools for many downstream tasks. They can be particularly effective when there are limited numbers of training samples for that downstream task. In the domain of camera physiological measurement there do not currently exist any public or published models trained on very large scale data. I believe that such a set of models would be very valuable for the community and contrastive learning could be one approach to creating them.

\subsection{Loss Functions.} In the design of supervised models the loss function used is important as it defines what will be optimized for in the learning process. In physiological sensing models there are typically two categories of loss function - waveform losses and metric losses. Waveform losses involve computing the error between a predicted and gold-standard physiologic (e.g., cardiac or breathing) waveform - which could typically be computed for every frame. Metric losses involve computing the error between a predicted metric, such as heart \emph{rate} or breathing \emph{rate}, and the gold-standard. This would apply for a window of time (e.g., at least one beat or breath). Because synchronization of data at the waveform level might be difficult (involving millisecond precision), often optimization is performed at the metric level for which synchronization need not be as precise. However, the relative frequency of feedback - once per time period versus once per frame - is lower which could impact the learning rate and the amount of training samples needed. Future work could compare these two to determine if optimization at the metric level leads to inferior recovered waveforms or conversely that it leads more precise downstream metrics.

\subsection{Meta-Learning}

Given the high individual variability in both visual appearance and physiological signals, personalization or customization of models becomes attractive. Several meta-learning techniques have been proposed for camera physiological measurement. Meta-RPPG~\cite{lee2020meta} was the first such approach which focuses on using transductive inference based meta-learning.  Liu et al.~\cite{liu2021metaphys} proposed a meta-learning framework built on top of the convolutional architecture previously presented \cite{liu2020multi}. They leveraged Model Agnostic Meta-Learning (MAML)~\cite{finn2017model} and tested both unsupervised and supervised model adaptation. The unsupervised method used pseudo PPG labels generated using POS~\cite{wang2017algorithmic}.  Meta-learning and model personalization should receive growing interest moving forward as it becomes possible to customize models more easily on-devices.

\subsection{Super Resolution and Video Enhancement}

Several methods have leveraged super resolution as a means of improving the extracted physiological waveforms, especially from low resolution input images. McDuff et al.~\cite{mcduff2018deep} showed that super resolution could help improve waveforms extracted from frames with us with resolution as low as 41 by 30 pixels. In this case, a neural super resolution was pairs with a traditional signal processing step extract the PPG signal. Yue et al.~\cite{yue2021deep} combined a neural super resolution step with a neural PPG extraction step to create a fully supervised example. As described above, neural approaches have been used to enhance videos before recovering the PPG signal. Spatio-temporal video enhancement can not only combat low spatial resolution but also the effects of video compression. The video enhancement network can be trained in a self supervised manner without requiring physiologic labels, and then a subsequent network fine tuned to recover the signal itself~\cite{yu2019remote}.

\section{Magnification and Visualization of Physiological Signals}
\label{sec:magnification}

\begin{wrapfigure}{L}{0.5\textwidth}
\centering
  \includegraphics[width=0.48\textwidth]{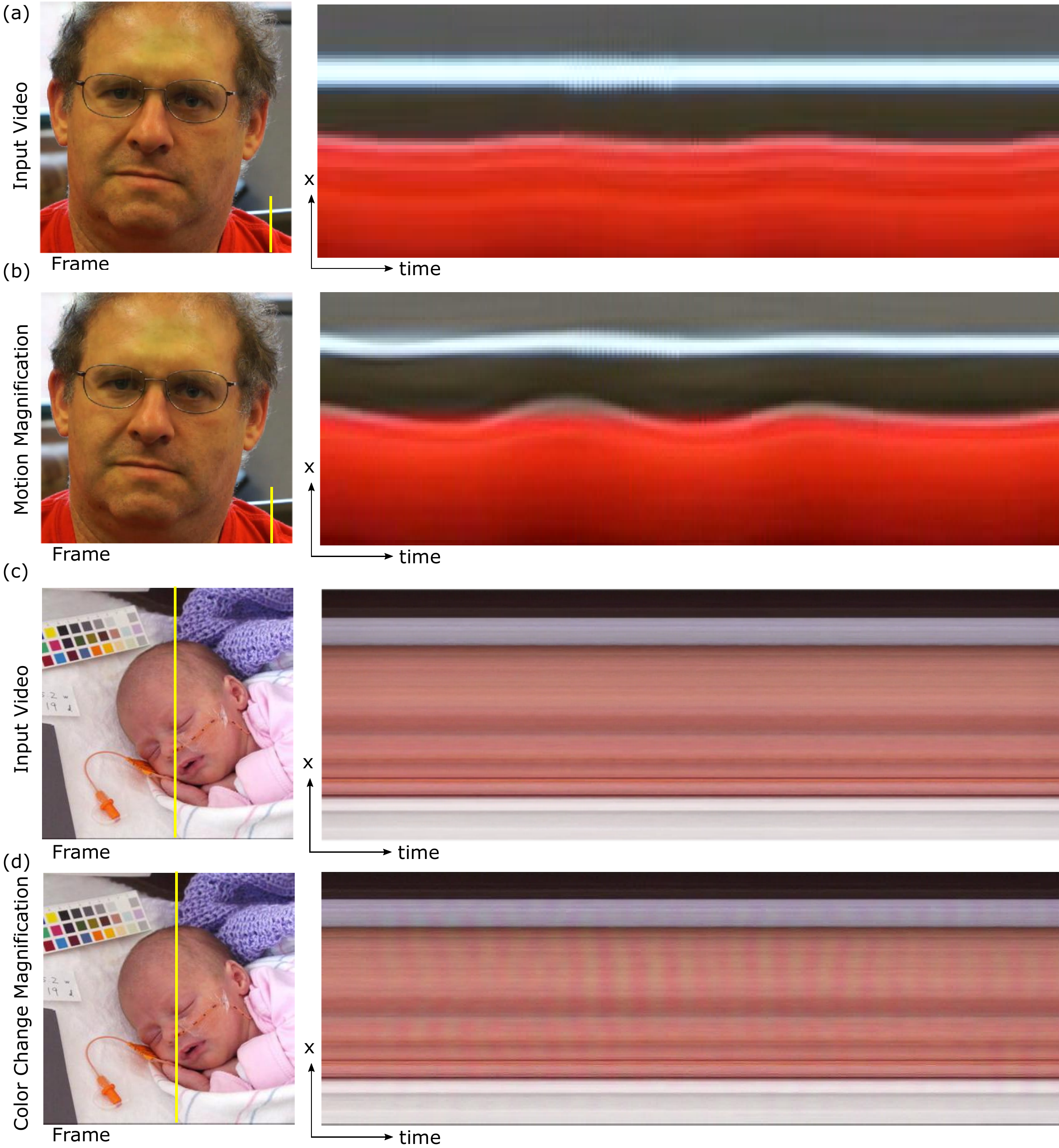}
  \caption{Examples of video magnification of physiological signals. Scan lines for motion (breathing) magnification method applied to the ``head" video and color change (pulse) magnification applied to the ``baby2" video from~\cite{wu2012eulerian}.}
  \label{fig:generalization}
\end{wrapfigure}

Camera physiological measurement enables certain opportunities that traditional sensors do not afford. Video magnification is an area of computational photography with the goal of magnifying changes of interest in a video. Magnification is helpful in cases where changes are subtle and difficult to see with the unaided eye. One application which has been used very frequently in this field is magnification of physiological changes in a video. Early video magnification methods used Lagrangian approaches which involve estimation of motion trajectories (e.g., the motion of the chest when someone is breathing) that are then amplified~\cite{liu2005motion,wang2006cartoon}. However, these approaches are often complex to implement in practice. The neat Eulerian video magnification (EVM) approach proposed by Wu et al.~\cite{wu2012eulerian} has had a significant impact on the field and raised the profile of video magnification as a whole. This method combines spatial decomposition with temporal filtering to reveal time varying signals without estimating motion trajectories. One draw back is that it uses linear magnification that only allows for relatively small magnifications at high spatial frequencies and cannot handle spatially variant magnification. 
To counter the limitation, Wadhwa et al.~\cite{wadhwa2013phase} proposed a non-linear phase-based approach, magnifying phase variations of a complex steerable pyramid over time. 
In general, the linear EVM technique is better at magnifying small color changes (i.e., more suitable for PPG), while the phase-based pipeline is better at magnifying subtle motions (i.e., more suitable for respiration, BCG or JVP). Both the EVM and the phase-EVM techniques rely on hand-crafted motion representations. To optimize the representation construction process, a supervised neural learning-based method \cite{oh2018learning} was proposed, which uses a convolutional network for frame encoding and decoding. With the learned motion representation, fewer ringing artifacts and better noise characteristics have been achieved. In preliminary work Pintea and van Gemert propose the use of phase-based motion representations in a learning framework that can be applied to the transference (or magnification) of motion~\cite{pintea2016making}.

One common problem with all the methods above is that they are limited to stationary subjects in which the physiological signal of interest is at another frequency (usually significantly faster) than other changes (e.g., body or camera motions), whereas many realistic physiological sensing applications would involve small changes of interest in the presence of large ones that might be at similar frequencies. For example, body motions might be at a similar frequency to the heart rate or breathing rate. After motion magnification, these large motions would result in large artifacts, and overwhelm any smaller variations. A couple of improvements have been proposed including a clever layer-based approach called DVMAG \cite{elgharib2015video}. By using matting, it can amplify only a specific region of interest (ROI) while maintaining the quality of nearby regions of the image. However, the approach relies on 2D warping (either affine or translation-only) to discount large motions, so it is only good at diminishing the impact of motions parallel to the camera plane and cannot deal with more complex 3D motions such as the human head rotation. The other method addressing large motion interference is video acceleration magnification (VAM) \cite{zhang2017video}. It assumes large motions to be linear on the temporal scale so that magnifying the motion acceleration via a second-order derivative filter will only affect small non-linear motions. However, the method will fail if the large motions have any non-linear components, and ideal linear motions are rare in real life, especially on living organisms.

Another problem with the previous motion magnification methods is that they use frequency properties to separate target signals from noise, so they typically require the frequency of interest to be known a priori for the best results and, as such, have at least three parameters (the frequency bounds and a magnification factor) that need to be tuned. If there are motion signals from different sources that are at similar frequencies (e.g., someone is breathing and turning their head), it is previously not possible to isolate the different signals.  Chen and McDuff presented a supervised learning approach that enables learning of source signals using gradient descent and then magnification using gradient ascent~\cite{chen2020deepmag}.

An example of the clinical utility of magnifying physiological signals using camera measurement was provided by Abnousi et al.~\cite{abnousi2019novel}. They used EVM to amplify videos of patients' necks and found that agreement between clinicians in the bedside assessment of the JVP was greater in the magnified condition compared to the unmagnified one. They argued that this technology could help expand the capabilities of telehealth systems.

\section{Clinical Applications and Validation}

\subsection{Neonatal Monitoring}

Neonates in intensive care require constant monitoring and are also active with clinical staff interacting with them regularly~\cite{chaichulee2019cardio}. The attachment of sensors can damage the skin and increase the risk of developing an infection or simply disrupt the sleep or comfort of an infant. Camera physiological measurement seems particularly well suited to this context. Numerous preliminary clinical validations studies have been conducted to access the readiness of these tools for monitoring neonates~\cite{aarts2013non,mestha2014towards,blanik2016remote,pereira2018noncontact,chaichulee2019cardio,villarroel2019non,gibson2019non,lorato2021towards}. Although the infants can and do move, they are relatively immobile (i.e., are laying down in a small incubator). Furthermore, illumination in a hospital environment can be controlled somewhat carefully. All in all, this is a promising application in which we might expect some degree of success. These initial validation studies have obtained promising results; however, further research is still needed to build confidence in the technology.
There are opportunities in this context to fuse signals from multiple sensors, such as pressure sensitive mats~\cite{lorato2021towards} which could offer additional benefits or help address some of the challenges of camera-based sensing, such as measurement when the body is obscured by blankets.

\subsection{Kidney Dialysis}

Validation of camera physiological measurements has also been performed in other clinical contexts. Specifically, Tarassenko et al.~\cite{tarassenko2014non} conducted experimentation to validate measurements on adult kidney dialysis patients. In this example, which was part of a larger scale clinical study, 46 patients had their vital signs monitored during 133 dialysis sessions. The advantages of camera sensing in this context are similar to those in the neonatal context. Removing the need for contact sensors could increase the comfort of the subjects helping them to sleep and move more easily.

\subsection{Telehealth}

One natural application of these technologies is in telehealth where platforms for video conferencing are used for remote patient care. The COVID-19 pandemic has highlighted the need for remote tools for measuring physiological states. With large numbers of telehealth visits being conducted over video conferencing platforms~\citep{annis_rapid_2020} there is still no scalable substitute for the measurements that would have traditionally recorded at a doctor's office. Therefore, computer visions tools for physiological measurement are attractive and becoming increasingly important~\citep{gawalko_european_2021, rohmetra_ai-enabled_2021}. However, to our knowledge, there are no published results from clinical validation studies using camera physiological measurement in this context. One challenge with these studies, unlike those performed in hospitals, is how to collect gold-standard sensor data while at the same time capturing videos that exhibit the natural variability that would be observed with patients joining from their home or another location.  It will certainly require a great amount of work to achieve this, but the potential benefits are significant.

\subsection{Sleep Monitoring}

Sleep studies are an important tool in diagnosing sleep disorders. Polysomnography (PSG) is the measurement of sleep via physiological sensing. However, the current PSG systems are cumbersome, disrupt sleep and require specialist equipment available only at sleep labs. Using NIR cameras several proof-of-concept studies have been performed demonstrating measurement of PPG~\cite{amelard2018non,vogels2018fully}, blood oxygen saturation~\cite{vogels2018fully,van2020camera} and breathing~\cite{van2020camera,scebba2020multispectral} (Scebba et al.~\cite{scebba2020multispectral} combined NIR and FIR cameras).
One study using camera vitals for sleep monitoring achieved pulse and respiratory rate detection within 2 beats/breaths per minute in over 90\% of samples and 4 percentage points error in blood oxygen saturation in 89\% of samples~\cite{van2020camera}. One sleeping disorder which  PSG can help to identify/diagnose is sleep apnea. Amelard et al.~\cite{amelard2018non} used a camera system to measure PPG and found pulse wave amplitude decreased during obstructed breathing and recovered after inhalation with a temporal phase delay. This early study provides encouraging evidence of the potential of video measurement during sleep. Camera systems are much easier to deploy and scale in homes than the equipement currently used for PSG.

\subsection{Health Screening}

Thermal imaging has been used for health screening at health clinics and airports for several years. Screening in this way can help limit the spread of infectious diseases and protect other people, including healthcare providers. Typically, these system measure body temperature. The limited availability of thermal cameras means that such systems cannot be deployed in every context. RGB and NIR imaging have some potential utility here; however, it is unclear if these sensor alone would be sufficient or whether they offer additional utility to thermal camera~\cite{sun2018noncontact}.

\section{Non-Clinical Applications}

\subsection{Baby Monitoring}

Outside of the clinical domain, consumer baby monitors are another set of products that can leverage camera physiological measurement. Similar arguments for camera measurement apply in consumer products as in the NICU applications (i.e., less disruption to the babies sleep and decreased risk of irritation or damage to the skin). Baby monitors that offer optical breathing measurement are already commercially available (e.g., MikuCare\footnote{https://mikucare.com/}). It is likely that the next generation of these devices will try to integrate heart rate measurement. Blood oxygen saturation would probably the next most likely signal. 
The role of physiological sensing for infants using consumer devices has been questioned, some argue that it could lead to increased anxiety about what these data mean\footnote{https://mashable.com/2017/02/18/raybaby-baby-breathing-monitor/}. As with all technologies, there is a need for user-centered design, demonstrating that the sensing and user interface solve a clear need for the consumer and minimise potential harms.

\subsection{Driver Monitoring}

Mitsubish Electric Research Labs (MERL)~\cite{nowara2018sparsePPG} and Toyota~\cite{gideon2021way} have both published research on camera physiological measurement in vehicles, this illustrates active interest from the automotive sector in these tools.  Signal processing~\cite{nowara2018sparsePPG} and neural~\cite{wu2019neural,gideon2021way} approaches have been proposed. 
In vehicle measurement could be used to help detect cardiac events and use this information to prevent accidents or for offering health monitoring as an attractive feature for customers. Demonstrations of camera physiological sensing can be found, but to our knowledge no vehicles currently on the market offer this facility.

\subsection{Biometrics}

Outside of the clinical realm or applications the focus on consumer health, Camera physiological measurement has received growing attention for detecting fake videos and verifying the ``liveness'' of a subject. Face verification tools could be fooled by a picture or a mask; however, it is very difficult to spoof subtle physiological changes in those cases. Researchers have leveraged this to detect deep fake videos~\cite{qi2020deeprhythm,ciftci2020fakecatcher} and propose anti-spoofing systems~\cite{liu20163d,liu2018learning}. For the latter, it is unclear if in practice these approaches would work effectively as the recovered signals can easily be corrupted. For example, it would not be hard to introduce a periodic change that is then picked up by the camera, and whether an imaging algorithm could determine a real versus fake period change is untested. Furthermore, the motion of a subject in front of a camera or heavy makeup may obstruct measurement of the PPG signal entirely making it appear as though a heat beat is not present when in fact it is.

\subsection{Affective Measurement}

The field of affective computing~\cite{picard2000affective} studies technology that measures, models, and responds to affective cues. Physiological signals contain information about autonomic nervous system activity. There are many other areas in which unobtrusive physiological sensing could help advance the vision of affective computing. Enabling measurement via ubiquitous devices increases opportunities to study affective signals in-situ and build systems that have the potential to be deployed at scale in the real-world.  
Two areas in which camera physiological measurement have been employed are: the detection of stress and cognitive load~\cite{bousefsaf2014remote,mcduff2016cogcam} and the measurement of responses to digital content~\cite{burzo2012towards,pham2015attentivelearner}. 

In order to build systems that respond to affective signals researchers have developed camera sensing of parameters closely related to sympathetic and parasympathetic nervous system activity. From cardiac signals heart rate variability or pulse rate variability has been used as a measure to quantify changes in cognitive load or stress~\cite{bousefsaf2014remote,mcduff2016cogcam}. The PPG signal contain several additional sources of information about autonomic nervous system activity. Blood volume pulse and vasomotion change in amplitude during stressful episodes~\cite{bousefsaf2014remote,mcduff2020non}.

\section{Challenges}

\subsection{Fairness}
\label{sec:fairness}

In camera physiological measurement appearance of the body or the environment is a key factor. Skin type, facial structure, facial hair, and clothing and other apparel can all affect the performance of measurement systems, as can lighting conditions. 

\textbf{Hardware.} Starting with the hardware, all cameras  are designed with certain operating criteria. Given the nature of the markets in which they are sold these cameras have often been optimized to capture lighter skin types more effectively then dark skin types. This can introduce an inherent bias in performance even if the algorithm and training data used do not.  Typically, sensitivity is greatest towards the middle of the camera's frequency range. Dark or very light skin types could be more likely to saturate the pixels and changes due to physiological variations may be lost.

\textbf{Data.} Almost all data sets for camera physiological measurement have been collected in Europe The United states or China (see Section~\ref{sec:datasets}). As such they predominantly contain images of lighter skin type participants. Furthermore, they generally feature younger people and often have a male bias. 
One challenge with constructing fair datasets in camera physiological sensing is that even the gold-standard contact devices can exhibit biases~\cite{bickler2005effects}. Evidence of biases in Sp0$_2$ measurement with skin type is prevalent, with three monitors tested over-estimating oxygen saturation in darker skin types in adults~\cite{bickler2005effects} and infants~\cite{vesoulis2021racial}. But other sensors (e.g., respiration, BCG, etc.) may also introduce biases. For example, chest straps frequently used as a gold-standard for measuring breathing may lead to different measurements on women than men.  Further characterization of camera \emph{and} gold-standard contact devices is needed to avoid errors from propogating, or worse compounding.

\textbf{Models.} The design of models for camera physiological measurement may also encode bias. This type of bias is often more difficult to detect. Several of the signal processing models described in Section~\ref{sec:computational_approaches} contain hard coded parameters but were evaluated primarily on datasets of light skin type subjects.
Some initial work has begun to characterize differences in performance of algorithms (both supervised and unsupervised) by different demographic and environmental parameters~\cite{addison2018video,nowara2020meta}. From preliminary research to clinical studies and the development of products I believe that this deserves greater attention. The development of balanced and representative datasets is one example of a significant contribution that could help towards this end. Meritable efforts towards this end have recently been published~\cite{chari2020diverse}.
Some of this work has also included novel methods for augmenting or simulating data to help address data imbalance amongst other things~\cite{ba2021overcoming}.
Dasari et al.~\cite{dasari2021evaluation} investigated the estimation biases of camera PPG methods across diverse demographics. As with previous work they observed similar biases as with contact-based devices and environmental conditions.  Chari et al.~\cite{chari2020diverse} proposed a physics-driven approach to help mitigate the effects of skin type on PPG measurement with encouraging results. I argue that innovations in hardware, better datasets and algorithmic contributions can all significantly improve the equitability in performance.

\subsection{Motion Tolerance}

The effects of subject motion have been among the most studied dynamics in camera physiological measurement. Understandably, much of the early work on focused on rigid, stationary subjects~\cite{wieringa2005contactless,humphreys2007noncontact,takano2007heart,verkruysse2008remote}. Subsequent studies allowed for limited naturalistic head motions~\cite{poh2010non}, but many experimental protocols still strictly limited the amount of motion during data collection~\cite{poh2010advancements,sun2012use}. Under these conditions and with reasonable image settings and illumination, recent methods will typically recover the underlying signals with high precision. For sleep measurement and in certain controlled contexts these assumptions may not be terribly unrealistic.  However, for others, such as consumer fitness applications (e.g., riding a static bike) they would be much too constrained. On this topic, work has examined pulse rate measurement during exercise on five different fitness devices~\cite{de2014improved}. This signal processing method was constrained by optical properties of the imager and the illumination source. As performance in constrained motion conditions began to saturate, researchers started to investigate algorithm performance under greater motion (both translational~\cite{hsu2017deep} and rotational~\cite{estepp2014recovering}). These approaches range from simple region of interest focused object tracking~\cite{yu2011motion} to projection~\cite{bartula2013camera} or signal separation~\cite{poh2010advancements} to more complex neural networks~\cite{chen2018deepphys}. Approaches for estimating the contribution of motion artifacts and correcting the PPG signal using adaptive filtering~\cite{cennini2010heart} or denoising networks~\cite{nowara2021benefit}
have also been explored. 

Several systematic and carefully controlled subject motion studies have been performed. Estepp et al.~\cite{estepp2014recovering} focused on rigid head rotations.  By combining data from multiple imagers and using blind source separation they were able to reduce the effects of rigid head motion artifact in the measurement of pulse rate~\cite{mcduff2017fusing}. 
In one way, multiple imagers simply add additional spatial redundancy, and methods have utilized this reduncancy in a single camera to reduce the impact of motion-induced artifacts
including translation, scaling, rotation, and talking~\cite{wang2014exploiting}. 
A comparable framework has been extended to include multiimagers in the infrared spectrum, as well~\cite{van2015motion}, and combining RGB and infrared imagers~\cite{gupta2016real,negishi2020contactless,scebba2020multispectral}.  Another approach, is to used motion information extracted via a body, head or face tracking system to filter or compensate for motion~\cite{chung2015high}.

Many of the aforementioned methods used signal processing approaches, without leveraging supervised learning. Neural models have proved highly effective at learning spatial and temporal information. Chen and McDuff~\cite{chen2020deepmag} illustrated this in the case of video magnification of PPG and breathing by showing how an algorithm could selectively magnify motions and color changes even in the presence of head motions at similar frequencies. All this being said, motion robustness should continue to be a focus in camera physiological measurement. Different types of motion are likely to be observed with different applications and so evaluation in the contexts of fitness and exercise, human computer interaction and video conferencing, baby monitoring and clinical care would all be very valuable.

\subsection{Ambient Lighting}

Ambient lighting conditions impact camera measurement in two primary ways, \emph{composition} and \emph{dynamics}. Composition of the ambient light can impact the performance of computational methods as absorption characteristics vary by frequency. The qualities and properties of constant illumination used for physiological measurement that are necessary to produce results of adequate quality have been explored~\cite{lin2017study}.  Brighter, green lighting tends to give the strongest improvement in PPG measurement. This is consistent with systematic analyses that have characterized the hemoglobin absorption spectra, although less is known about how light composition might impact motion-based analysis (i.e., breathing or BCG measurement). 
From a practical perspective, light composition is not only tied to the absorption or reflectance properties of the body, but also the image quality.  As I shall discuss in the following subsection, lens aperture and focus, sensor sensitivity (ISO), individual frame exposure time (integration, shutter speed) and other image settings will also impact performance. 

If the intensity, position or direction of lighting changes dynamically it will typically introduce relatively large pixel changes compared to those resulting from physiological processes. Where ambient lighting conditions can be controlled and/or held relatively constant it can be extremely advantageous. There are several cases in which this might be true (e.g., an incubator in a hospital or in a gym) and cases in which this almost certainly will not be true (e.g., driving, etc.)  Changes in illumination intensity affect absolute magnitude of camera measured PPG waveforms~\cite{sun2012use}, which can in turn impact measures of pulse wave amplitude and blood composition. However, it is still true that the effects are still largely uncharacterized for many physiological signals~\cite{mcduff2015survey}. 
Several computational methods have been proposed to help combat lighting effects. Li et al.~\cite{li2014remote} used an adaptive filtering approach, with an isolated background region of interest serving as the input noise reference signal, to compensate for background illumination. Nowara et al.~\cite{nowara2021benefit} used a similar concept, leveraging an inverse of the PPG attention mask as the background region. These methods both provided an overall reduction in heart rate error. 
However, in the case of Li et al.~\cite{li2014remote} an ablation study of the components of their multi-stage processing approach (region of interest detection, illumination variation correction, data pruning, and temporal filtering) was not made available in order to sufficiently determine the effectiveness of any single stage. Neither did Nowara et al.~\cite{nowara2021benefit} identify if the inverse attention was primarily addressing illumination changes, body motions or other sources of noise. 
Amelard et al.~\cite{amelard2015illumination} presented results using  a temporally-coded light source and synchronized camera for PPG measurement in dynamic ambient lighting conditions. Novel hardware presents some interesting opportunities for combating illumination; however, understandably most work focuses on ``off-the-shelf'' cameras, due to the lower technical barrier and far greater availability of those devices.

To the best of my knowledge no research in camera physiological measurement has made use of computational color space calibration and white balancing methods~\cite{nguyen2014training}. Priors on skin color can help correct color inconsistencies in images~\cite{bianco2014adaptive} and it may be possible then that the inverse could be true, priors on scene color could be used to correct skin pixel color inconsistencies with or across videos.
There are also methods for relighting faces, using as little as a single frame~\cite{sun2019single}.  Both of these approaches could be helpful for relighting/altering color profiles at test time to help a model perform more accurately, or augmenting a training set to help build models that generalize better. Of course, these hypotheses need rigorous validation, but there certain appear to be many tools in computer vision, graphics and computational photography that could aid in camera physiological measurement.

\subsection{Imaging Settings}

While cameras are ubiquitous, they vary considerably in specifications. This is one reason obtaining regulatory approval for camera-based solutions can be challenging. Determining the optimal qualities of an image sensor and characterizing how sensitive measurements are to changes in there parameters is very valuable. These parameters include sensor type (e.g. CCD, CMOS), color filter array (e.g. Bayer, Fovenon X3, and RGBE), number and specification of frequency bands, bit depth, imager size, and number of pixels. Beyond the image sensor, there are other hardware considerations, such as lens type and quality, spectral properties of the illumination source, and image aperture/shutter speed/ISO. All of these parameters will affect the overall content of any acquired image.  Then there are software properties or controls, some of which may be constrained by the hardware and others by the bandwidth or storage capabilities. These include, the resolution of the video frames, the frame rate at which the video is captured, whether white balancing, auto focus, or brightness controls are enabled and dynamically changing during video capture. Given the myriad of combinations here, it is understandable that it is difficult to precisely characterize the impact of each. Needless to say, sensor quality, resolution and frame rate all play a particular role. It may be possible to use intuition to help guide some of these judgements, for example shutter times should avoid pixel saturation~\cite{van2021remote}.

It is also important to consider how the apparatus/equipment set-up and context with impact signal quality for a given hardware and software configuration. 
The distance of the body region of interest from the camera will pack the pixel density and the additional contextual information that might be available from the image. Helpfully, some datasets have characterized the face ROI pixel density~\cite{kopeliovich2019color}, we recommend that future datasets do similarly. Placing a camera very close to the body might lead to a higher number of pixels containing the signal of interest, but also potentially mean that information about other related signals, or context (e.g., body motions/activities) is lost.

Some studies exist on the comparison of multiple imagers running in parallel during data collection (e.g.~\cite{sun2012use,niu2018vipl}) and offer some confidence that signal recovery can be robust over widely varying imager properties. Studies of image size (pixel density) and frame rate in single-imager~\cite{sun2012noncontact} and multi-imager~\cite{estepp2014recovering,gupta2016real,scebba2020multispectral} sensor designs have shown that, as expected, these parameters do impact the performance of PPG measurement. Breathing measurement is impacted more significantly by pixel density, which is why Chen and McDuff used a higher resolution for breathing model magnification than PPG magnification~\cite{chen2020deepmag}.

As with many of the topics discussed in this section, a great contribution would be the creation and standardization of an explicit benchmark test, and related metrics, that could be performed with a variety of imagers to better understand and compare results across studies and methods. The VIPL dataset~\cite{niu2018vipl} is the closest example of such a dataset (and will be described in detail in Section~\ref{sec:datasets}).

\subsection{Video Compression}

Video compression algorithms are designed to reduce the total number of bits needed to represent a video sequence. These algorithms have been traditionally designed to preserve video quality, characterized by scores related to human perceptual quality, for example minimizing motion artifacts and loss of clarity. Compression algorithms have not been designed directly, or indeed indirectly in most cases, for preserving physiological information with a video. 
Compression can impact measurements that rely on motion (e.g., breathing) less than those that rely on color changes (e.g., PPG)~\cite{nowara2021systematic}, but will to some degree impact both. The subtle changes in pixel values that are used to recover the PPG signal are often imperceptible to the eye and these small temporal variations are often removed by compression algorithms to reduce the overall bitrate. Previous work in systematically analyzed the impact a video compression on PPG measurement~\cite{mcduff2017compression,rapczynski2019effects} found a linear decrease in the PPG signal-to-noise ratio (SNR) with increasing constant rate compression factors. However, Rapczynski et al.~\cite{rapczynski2019effects} observed that performance of HR estimation was less sensitive to decreases in resolution and color subsampling, both of which can be used to reduce video bitrates. 

There are other ways to reduce the impact of compression on physiological signals within a video, for example by training a model on videos at the same compression level~\cite{nowara2021systematic}. Supervised models can learn to reverse or ignore compression artifacts to some degree. Given video compression is necessary for many applications (i.e., in cloud-based teleconferencing systems) this insight may prove useful. It would certainly be impractical with current bandwidth limits to stream raw video at scale. Recovering the signals from heavily compressed videos is something that deserves further attention. Yu et al.~\cite{yu2019remote} designed a video enhancement model that could serve this purpose and be trained in a self supervised manner. 
Datasets with varying levels of video compression are somewhat easy to create and I argue that standard versions of all public datasets could and should be created with multiple video compression levels so that researchers can report results across different compression rate factors. Zhao et al.~\cite{zhao2018novel} proposed such a benchmark dataset; however, access to that data is unclear.

\section*{Ethics and Privacy Implications}
\label{sec:broader_impacts}

The many positive applications of the measurement of physiological signals using cameras illustrates that this technology has great potential. However, there are very important risks to consider and potential mitigations that can be put in place to minimize the impact of these risks. Cameras are an unobtrusive and ubiquitous form of sensor, that are used for surveillance at scale. Using similar methods to those described for monitoring patients in intensive care, a ``bad actor'' could employ these tools for surveilling people. Cameras could be used to measure personal physiological information without the knowledge of the subject. Military or law enforcement bodies may try to apply this in an attempt to detect individuals who appear ``nervous'' via signals such as an elevated heart rate or irregular breathing. Or an employer may surreptitiously screen prospective employees for health conditions without their knowledge during an interview (e.g., heart arrhythmias or high blood pressure). Some may attempt to justify these applications by claiming that monitoring could also be used to screen for symptoms of a virus during a pandemic or to promote public safety.

There are several reasons that this would be irresponsible and harmful. First, there is little evidence that physiological signals would provide enough information, without additional context, for determining emotional states or job eligibility. Second, camera physiological measurement still requires significant validation in real-world settings and it is unlikely that the current state-of-the-art camera physiological measurement systems would be accurate enough in these context. As described in Section~\ref{sec:fairness}, there is evidence that they currently do not perform with equal accuracy across people of all appearances and in all contexts.  The populations that are subject to the worst accuracy might also be those that are already subject to disproportionate targeting and systematic negative biases~\cite{garvie2016perpetual}. Many of these issues have been discussed in the context of facial recognition; but parallels can be drawn with physiological measurement. Third, there are many possible negative social outcomes that might result even if measurement was ``accurate''. Normalizing covert surveillance of this kind can be dangerous and counterproductive. 

As with any new technology, it is important to consider how camera physiological measurement could be applied in a negligent or irresponsible manner whether by individuals or organizations.  Application without sufficient forethought for the implications could undermine the positive applications of these methods and increase the likelihood that the public will mistrust the tools. 

These applications would set a very dangerous precedent and would probably be illegal. Just as is the case with traditional contact sensors, it must be made very transparent when camera-based physiological measurement is being used and subjects should be required to consent data is collected. There should be no penalty for individuals who decline to be measured. Ubiquitous sensing offers the ability to measure signals in more contexts, but that does not mean that this should necessarily be acceptable. Just because cameras may be able to measure these signals in new context, or with less effort, it does not mean they should be subject to any less regulation than existing sensors.

While far from a solution to the challenges described above, researchers have proposed innovative methods for removing physiological information from videos~\cite{chen2017eliminating} and ``blocking'' video-based measurement~\cite{mcduff2018inphysible}.  There are also instances of more generic computer vision jamming systems~\cite{harvey2012cvdazzle,yamada2013privacy,wilber2016can} that could apply in the context of camera physiological measurement. However, we should recognize that these solutions often put the onus on the subject to opt-out and could be very inconvenient and stigmatizing. The emphasis should be on opt-in systems that are used in well validated and regulated contexts.

\section{Software}

In this section I highlight some of the repositories of open source code for camera physiological sensing. Unlike other domains in machine learning there are relatively few complete repositories containing implementations of baseline methods. The research community would do well to address this.

\textbf{MATLAB.} For signal processing analysis MATLAB has often been a popular language for implementation.  McDuff and Blackford~\cite{mcduff2019iphys}\footnote{\url{https://github.com/danmcduff/iphys-toolbox}} implemented a set of source separation methods (Green, ICA, CHROM, POS) in MATLAB and Pilz~\cite{pilz2019vector} published the PPGI-Toolbox\footnote{\url{https://github.com/partofthestars/PPGI-Toolbox}} containing implementations of Green, SSR, POS, Local Group Invariance (LGI), Diffusion Process (DP) and Riemannian-PPGI (SPH) models.

\textbf{Python.} 
Increasingly, Python is becoming more popular as a language for developing camera physiological measurement methods.  There are several implementation of the popular signal processing methods: Bob.rrpg.base\footnote{\url{https://pypi.org/project/bob.rppg.base/}} includes implementations of CHROM, SSR and Li et al.~\cite{li2014remote} and Boccignone et al.~\cite{Boccignone2020} released code for Green, CHROM, ICA, LGI, PBV, PCA, POS, SSR.  Several published papers have included links to code; however, often this is only inference code and not training code. 

To date, there are very few code bases that provide implementations of multiple supervised neural models, despite these being the best performing methods. Researchers have released code for their own methods, often accompanying papers; however, a unified code base or toolbox is not available.

\afterpage{%
\clearpage
\begin{landscape}
\begin{table}
	\caption{Summary of Imaging Datasets}
	\label{tab:methods}
	\centering
	\small
	\setlength\tabcolsep{6pt} 
	\begin{tabular}{lp{3cm}p{1cm}p{1cm}p{3.5cm}p{2cm}p{3.5cm}}
	\toprule
	Dataset & Context & Subjects & Videos & Imaging & Gold Standard & Other Comments \\
	\hline \hline
	MAHNOB-HCI~\cite{soleymani2011multimodal} & Implicit media tagging & 27 & 527 & \makecell{Resolution: 780x580 \\ Frame Rate: 61Hz} & ECG, EEG, Breathing & Videos are quire heavily compressed. \\
	
	BP4D~\cite{zhang2016multimodal} & Multimodal affect analysis & 140 & 1400 & \makecell{Resolution: 1040x1392 \\ Frame Rate: 24Hz} & Blood Pressure Wave & No PPG gold-standard but continuous fingertip blood pressure.  \\
	
	VIPL-HR~\cite{niu2018vipl} & Camera physiology &  107 & 3130  & \makecell{Resolution: 960x720/640x480 \\ Frame Rate: 25Hz/20Hz} & PPG, HR, SpO$_2$ & 2378 RGB and 752 NIR \\ %
	
	COHFACE~\cite{heusch2017reproducible} & Camera physiology & 40 & 160 & \makecell{Resolution: 640x480 \\ Frame Rate: 20Hz} & PPG & \\
	
	UBFC-RPPG~\cite{bobbia2019unsupervised} &  Camera physiology & 42 & 42 & \makecell{Resolution: 640x480 \\ Frame Rate: 30Hz} & PPG, PR \\
	
	UBFC-PHYS~\cite{meziatisabour2021ubfc} &  Camera physiology & 56 & 168 & \makecell{Resolution: 1024×1024 \\ Frame Rate: 35} & PPG, EDA & Contact measurements were obtained using a wrist worn device and therefore the signal quality is variable. \\
	
	RICE CameraHRV~\cite{pai2018camerahrv} &  Camera physiology &  12 & 60 & \makecell{Resolution: 1920×1200 \\ Frame Rate: 30Hz} & PPG \\ 
	
	MR-NIRP~\cite{nowara2018sparsePPG} &  Camera physiology &  18  & 37 & \makecell{Resolution: 640x640  \\ Frame Rate: 30Hz} & PPG & Contains videos recorded during driving. \\
	
	PURE~\cite{stricker2014non} & Camera physiology &  10 &  59  & \makecell{Resolution: 640x480 \\ Frame Rate: 60Hz} & PPG, SpO$_2$\\
	
	rPPG~\cite{kopeliovich2019color} &  Camera physiology & 8 & 52 & \makecell{Resolution:  1920x1080 / 640x480 \\ Frame Rate: 15Hz } & PR, SpO$_2$ \\
	
	OBF~\cite{li2018obf} &  Camera physiology &  106 & 212 & \makecell{Resolution: 1920x1080 (RGB) \\ Frame Rate: 60Hz} & PPG, ECG, RR & Contains a small subset of subjects with atrial fibrillation.  \\
	
	PFF~\cite{hsu2017deep} &  Camera physiology & 13 & 85 & \makecell{Resolution: 1280x720 \\ Frame Rate: 50Hz} & PR \\
	
	CMU~\cite{dasari2021evaluation} &  Camera physiology &  140 & 140 & \makecell{Resolution: 25x25 \\ Frame Rate: 15Hz} & PR & Videos are anonymized and only show skin regions on the forehead and cheeks, not the full face. \\
	
    \bottomrule
  \end{tabular}
\end{table}
\end{landscape}
\clearpage
}

\section{Data}

\subsection{Public Datasets}
\label{sec:datasets}

Public datasets serve two important purposes for the research community. First, they provide access to data to researchers who many not have the means to collect their own, lowering the bar to entry.  Second, the provide a transparent testing set to fairly compare computational methods and set benchmarks. Descriptions of benchmark datasets should include details of the imaging device, lighting and participant demographic information. In addition to videos and gold-standard contact measurements.

The first two datasets on this list have been commonly used for benchmarking camera physiological measurement, but were not collected explicitly for that purpose:

\textbf{MAHNOB-HCI.}~\cite{soleymani2011multimodal}\footnote{\url{https://mahnob-db.eu/hci-tagging/}} The MAHNOB-HCI dataset was originally collected for the purposes of creating systems for implicit tagging of multimedia content. Videos of 27 participants (15 women, 12 men) were collected while they were wearing an ECG sensor. This was one of the earliest public datasets which included videos and time synchronized physiological groundtruth. One limitation of this data set is the heavy video compression which means that physiological information in the videos is somewhat attenuated. Videos were recorded at a resolution of 780x580 and 61Hz. Most analyses~\cite{li2014remote,chen2018deepphys}, use a 30-second clip (frames from 306 through 2135) from 527 video sequences.

\textbf{BP4D+ and MMSE-HR.}~\cite{zhang2016multimodal}\footnote{\url{http://www.cs.binghamton.edu/~lijun/Research/3DFE/3DFE_Analysis.html}} The BP4D+ data set is a multimodal data set containing time synchronized 3D 2D thermal and physiological recordings. This large data set contains videos of 140 subjects and ten emotional sitting tasks. The videos astorg in relatively uncompressed format add the data set contains a relatively broad range of ages 18 to 66 and ethnic or racial diversity. Furthermore unlike many other datasets was there contains a majority female. of note is that this data set does not include either PPG or ECT gold standard measures but rather contains pulse pressure waves as measured fire fingercuff. The post pressure wave is similar to but different in morphology to the PPG signal. RGB videos were recorded at a resolution of 1040x1392 (Note: this is portrait) and 24 Hz.
\newline \newline 
The following datasets were collected for the explicit purposes of developing and benchmarking camera physiological measurement methods:

\textbf{VIPL-HR.}~\cite{niu2018vipl}\footnote{\url{https://vipl.ict.ac.cn/view_database.php?id=15}} VIPL-HR is the largest multimodal data set with videos and time synchronized physiological recordings it contains 2378 RGB or visible light videos and 752 near infrared videos of 107 subjects. Gold-standard PPG, heart rate and SpO$_2$ were recorded.
Videos were recorded with three RGB cameras and one NIR camera: i) an RGB Logitech C310 at resolution 960×720 and 25 Hz, ii) a RealSense F200 NIR camera at resolution 640×480 and RGB camera at 1920×1080, both 30 Hz, iii) an RGB HUAWEI P9 at resolution 1920×1080 and 30 Hz. 

\textbf{COHFACE.}~\cite{heusch2017reproducible}\footnote{\url{https://www.idiap.ch/en/dataset/cohface}} The COHFACE dataset contains RGB video recordings synchronized with cardiac (PPG) and respiratory signals. The dataset includes 160 one-minute long video sequences of 40 subjects (12 females and 28 males). The video sequences have been recorded with a Logitech HD C525 at a resolution of 640x480 pixels and a frame-rate of 20Hz. Gold-standard measurements were acquired using the Thought Technologies BioGraph Infiniti system.

\textbf{UBFC-RPPG}~\cite{bobbia2019unsupervised}\footnote{\url{https://sites.google.com/view/ybenezeth/ubfcrppg}}
The UBFC-RPPG RGB video dataset, collected with a Logitech C920 HD Pro at 30Hz with a resolution of 640x480 in uncompressed 8-bit RGB format. A CMS50E transmissive pulse oximeter was used to obtain the gold-standard PPG data. During the recording, the subjects were seated one meter from the camera. All experiments are conducted indoors with a mixture of sunlight and indoor illumination. 

\textbf{UBFC-PHYS.}~\cite{meziatisabour2021ubfc}\footnote{\url{https://sites.google.com/view/ybenezeth/ubfc-phys}} UBFC-PHYS is another public multimodal dataset with RGB videos, in which 56 subjects (46 women and 10 men) participated in Trier Social Stress Test (TSST) inspired experiment. Three tasks (rest, speech and arithmetic) were completed by each subject resulting in 168 videos. Gold-BVP and EDA measurements were collected via a wristband (Empatica E4). Before and after the experiment, participants completed a form to calculate their self-reported anxiety scores. The video recordings were at resolution 1024x1024 and 35Hz. 

\textbf{Rice CameraHRV.}~\cite{pai2018camerahrv}\footnote{\url{https://sh.rice.edu/camerahrv/}}
The Rice CameraHRV consists of activities with complex facial movement, containing video recordings of 12 subjects (8 male, six female) during stationary, reading, talking, video watching and deep breathing tasks (total of 60 recordings). Each video is 2 minutes in duration. Gold-standard PPG data were collected using an FDA approved pulse oximeter. The camera recordings were made with a Blackfly BFLY-U3-23S6C (Point Grey Research) with Sony IMX249 sensor. Frames were captured at a resolution of 1920x1200 and 30Hz. 

\textbf{MERL-Rice NIR Pulse (MR-NIRP).}~\cite{nowara2018sparsePPG}\footnote{\url{ftp://merl.com/pub/tmarks/MR_NIRP_dataset/README_MERL-Rice_NIR_Pulse_dataset.pdf}}
The MR-NIRP dataset contains recordings (19) of drivers in a cockpit driving around a city and recordings (18) stationary in a garage. Each video recorded in the garage is two minutes in duration and those recorded while driving are 2-5 minutes long. The 18 (16 male, two female) subjects were healthy, aged 25–60 years. Four of the subjects were recorded at night and 14 during the day. Recordings were made with NIR (Point Grey Grasshopper GS3-U3-41C6NIR-C) and RGB (FLIR Grasshopper3 GS3-PGE23S6C-C) cameras mounted on the dashboard in front of the subject. The NIR camera was fitted with a 940 nm hard-coated optical density bandpass filter from Edmund Optics with a 10 nm passband. Frames were captured at a resolution of 640x640 and 30Hz (no gamma correction and with fixed exposure).
Gold-standard PPG data were recorded with a CMS 50D+ finger pulse oximeter at 60Hz.

\textbf{PURE.}~\cite{stricker2014non} The PURE datasets contains recordings of 10 subjects (8 male, 2 female) each during six tasks.  The videos were captured with an RGB eco274CVGE camera (SVS-Vistek GmbH) at a resolution of 640x480 and 60 Hz.  The subjects were seated in front of the camera at an average distance of 1.1 meters and lit from the front with ambient natural light through a window. Gold-standard measures of PPG and SpO$_2$ were collected with a pulox CMS50E attached to the finger.
The six tasks were described a follows: i) The subject was seated, stationary and looking directly into the camera. ii) The subject was asked to talk while avoiding additional head motion. iii) The participant moved their head in a horizontal translational manner at an average speed proportional to the size of the face within the video. iv) Similar to the previous task with twice the velocity. v) Subjects were asked to orient their head towards targets placed in an arc around the camera in a predefined sequence. The motions were designed to be random, and not periodic (approx. 20° roations). vi) Similar to the previous task with larger head rotations (approx. 35° rotations).

\textbf{rPPG}~\cite{kopeliovich2019color}\footnote{\url{https://osf.io/fdrbh/wiki/home/}}
The rPPG dataset includes 52 recording from three RGB cameras: a Logitech C920 webcam at resolution 1920×1080 (WMV2 video codec), a Microsoft VX800 webcam at resolution 640 × 480 (WMV3 video codec), and a Lenovo B590 laptop integrated webcam at resolution 640×
480 pixels (WMV3 video codec). All recordings were 24-bit depth (3x 8-bit per channel) at 15 Hz. The duration of the recordings was between 60 and 80 seconds. Between 2 and 14 videos were recorded for eight healthy subjects (7 male, 1 female, 24 to 37 years). Primary illumination was ambient daylight and indoor lighting.
Subjects were seated 0.5-0.7 m from the camera. 
Gold-standard PR measures were collected via a Choicemmed MD300C318 pulse oximeter. Participants completed a combination of stationary and head motion tasks. In the motion tasks, subjects rotated their head from right to left (with 120° amplitude), from up to down (with 100° amplitude). Subject was also asked to speak and change facial expressions. 

\textbf{OBF.}~\cite{li2018obf} The Oulu Bio-Face (OBF) database includes facial videos recorded from healthy subjects and from patients with atrial fibrillation. Recordings were made with an RGB and NIR camera. The subjects were seated one meter from the cameras. Two light sources were placed either side of the cameras and illuminated the face at 45 degree angle from a distance of 1.5 meters.
According to their published work the authors plan to make this dataset publicly available; however, I was unable to find information about how to access it at the time of writing.

\textbf{PFF.}~\cite{hsu2017deep}\footnote{\url{https://github.com/AvLab-CV/Pulse-From-Face-Database}} The PPG From Face (PPF) database includes facial videos of 13 subjects each during five tasks (65 videos total). Each video is 3 minutes, recorded with resolution 1280x720 at 50 Hz. Gold-standard PR was collected via two Mio Alpha II wrist heart rate monitors (the average PR of the two readings is used).  The subjects were seated in front of the camera at a distance of 0.5 meters. 
The five tasks were: 1) The subject was seated stationary with fluorescent illumination. 2) The subject moved their head/body in a horizontal translational motion (right and left) with a frequency between 0.2-0.5 Hz. Flourescent lights were on. 3) The subject was seated stationary with ambient illumination primarily from windows and a computer monitor. 4) The same as task 2 with ambient illumination primarily from windows and a computer monitor. 5) The same illumination condition as Task 1, each subject was riding on an exercise bike at a constant speed.

\textbf{CMU.}~\cite{dasari2021evaluation}\footnote{\url{https://github.com/AiPEX-Lab/rppg_biases}}  A new CMU rPPG dataset contains videos recorded from 140 subjects subjects in India (44) and Sierra Leone (96). Three deidentified videos were generated from each face video, one each of the forehead, left cheek and right cheek. A rectangular region of resolution 60x30 of the forehead, a square region of resolution 25x25 pixels of the left cheek and a square region of 25x25 pixels of the right cheek. Videos were recorded at 15 Hz.

\subsection{Synthetics and Data Augmentation}

Labeled data is a limiting factor in almost all areas of computer vision and machine learning. Even large datasets can often suffer from selection bias and a lack of diversity. Although there are no easy solutions to these problems, the are methods for alleviating the problem: 1) data augmentation, 2) data simulation of synthesis.

Several recent papers have proposed methods of data augmentation by creating videos with modified or augmented physiological patterns. Both Gideon and Stent~\cite{gideon2021way} and Niu et al.~\cite{niu2019robust} use resampling to augment the frequency, of temporal, information in videos.  While the former example performed augmentation in the video space, the latter example performed the data on their spatial-temporal feature space arguing that it preserves the HR distributions.
Specifically, to address the issue of the lack of representation of skin type in camera physiology dataset, Ba et al.~\cite{ba2021overcoming} translate real videos from light-skinned subjects to dark skin subjects while being careful to preserve the PPG signals. A neural generator is used to simulate changes in melanin, or skin tone. This approach does not simulate other changes in appearance that might also be correlated with skin type.
Nowara et al.~\cite{nowara2021combining} use video magnification (see Section~\ref{sec:magnification} for augmenting the spatial appearance of videos and the temporal magnitude of changes in pixels. These augmentations help in the learning process, ultimately leading to the model learning better representations.

Recent work has proposed using simulations to create synthetic data for training camera physiological measurement algorithms. This can take two forms, statistical generation of videos using machine learning techniques or simulation using parameterized computer graphics models. Tsou et al.~\cite{tsou2020multi} used the former approach, leveraging neural models for video generation from a source image and a target PPG signal. Generative modeling definitely offers many opportunities in physiological measurement~\cite{song2021pulsegan,lu2021dual}.
Computer graphics can provide a way to create high fidelity videos of the human body with augmented motions and skin subsurface scattering that simulate cardiac and respiratory processes~\cite{mcduff2020advancing}. Synthetics pipelines have the advantage of allowing simulation of many different combinations of appearance types, contexts and physiological states example high heart rates or arrhythmia states for which it may be difficult to create to gather examples in a lab. Research has shown that greater and greater numbers of avatars in a synthetic training set can continue to boost performance up to a point~\cite{mcduff2021synthetic}. However this is early work and there remains a ``sim-to-real'' gap in performance of these systems, models trained purely on synthetic data do not generalize perfectly to real videos. Furthermore, these synthetics pipelines are typically expensive to construct and therefore there may be limited access to them.

\section{Conclusion}

I have presented a survey of camera physiological measurement methods, these techniques have huge potential to improve the noninvasive measurement and assessment vital signs. Camera technology and computational methods have advanced dramatically in the past 20 years benefiting from advancements in optics, machine learning and computer vision. With applications from consumer fitness to telehealth to neonatal monitoring to security and affective computing there are many opportunities for these methods to have impact in the next 20 years. However, there are significant challenges that will need to be addressed in order to realize that vision. These include but are not limited to addressing: unfair and inequitable performance, environmental robustness, the current lack of clinical validation and privacy and ethical concerns. The ethical challenges associated with camera sensing should not be disregarded or treated lightly. While there is a role for technological solutions that make it easier to remove physiological information from video, it is much more important to make sure these technologies are always designed in an opt-in manner.

\begin{acks}
I would like to thank all my collaborators who have contributed to work on camera physiological measurement over the past 10 years, Ming-Zher Poh, Rosalind Picard, Javier Hernandez, Sarah Gontarek, Ethan Blackford, Justin Estepp, Izumi Nishidate, Vincent Chen, Xin Liu, Ewa Nowara and Brian Hill. I would also like to thank Wenjin Wang and Sander Stuijk for co-organizing the Computer Vision for Physiological Measurement (CVPM) workshops which have helped to consolidate the research community around these methods, from which a lot of this work came.
\end{acks}

\bibliographystyle{ACM-Reference-Format}
\bibliography{sample-base}

\end{document}